\documentclass[10pt,twocolumn,letterpaper]{article}

\usepackage{cvpr}
\usepackage{times}
\usepackage{epsfig}
\usepackage{graphicx}
\usepackage{amsmath}
\usepackage{amssymb}
\usepackage{dsfont}
\usepackage{siunitx}
\usepackage{subfigure}
\usepackage{multirow}
\usepackage{caption}
\usepackage{xcolor}
\graphicspath{{./images/}}

\def\argmax{\mathop{\rm argmax}	}

\def\1{\mathds{1}}

\def\eg{{e.g.$\;$}}

\newcommand{\myparagraph}[1]{\vspace{6pt}\noindent{\bf #1}}

\newcommand{\modif}[1]{\textcolor{blue}{#1}}

\usepackage[breaklinks=true,bookmarks=false]{hyperref}

\cvprfinalcopy 

\def\cvprPaperID{1896} 

\begin{document}

\title{Zero-Shot Learning - The Good, the Bad and the Ugly}

 \author{
 Yongqin Xian$^{1}$ \hspace{4mm} Bernt Schiele$^{1}$ \hspace{4mm} Zeynep Akata$^{1,2}$\vspace{4mm} \\ 
  \begin{tabular}{cc}
  $^{1}$Max Planck Institute for Informatics & $^{2}$Amsterdam Machine Learning Lab \\ Saarland Informatics Campus & University of Amsterdam 
 \end{tabular}
 }

\newcommand{\edit}[1]{\textcolor{red}{\textbf{Modified: #1}}}
\newcommand{\add}[1]{\textcolor{red}{\textbf{Added: #1}}}
\newcommand{\toAdd}[1]{\textcolor{red}{\textbf{To add #1?}}}
\maketitle

\begin{abstract}
Due to the importance of zero-shot learning, the number of proposed approaches has increased steadily recently. We argue that it is time to take a step back and to analyze the status quo of the area. The purpose of this paper is three-fold. First, given the fact that there is no agreed upon zero-shot learning benchmark, we first define a new benchmark by unifying both the evaluation protocols and data splits. This is an important contribution as published results are often not comparable and sometimes even flawed due to, e.g. pre-training on zero-shot test classes. Second, we compare and analyze a significant number of the state-of-the-art methods in depth, both in the classic zero-shot setting but also in the more realistic generalized zero-shot setting. Finally, we discuss limitations of the current status of the area which can be taken as a basis for advancing it. 
\end{abstract}

\section{Introduction}
\label{sec:intro}

Zero-shot learning aims to recognize objects whose instances may not have been seen during training~\cite{HEEY15,LNH13,LEB08,RSS11,YA10}. The number of new zero-shot learning methods proposed every year has been increasing rapidly, i.e. the good aspects as our title suggests. Although each new method has been shown to make progress over the previous one, it is difficult to quantify this progress without an established evaluation protocol, i.e. the bad aspects. In fact, the quest for improving numbers has lead to even flawed evaluation protocols, i.e. the ugly aspects. Therefore, in this work, we propose to extensively evaluate a significant number of recent zero-shot learning methods in depth on several small to large-scale datasets using the same evaluation protocol both in zero-shot, i.e. training and test classes are disjoint, and the more realistic generalized zero-shot learning settings, i.e. training classes are present at test time. 

We benchmark and systematically evaluate zero-shot learning w.r.t. three aspects, i.e. methods, datasets and evaluation protocol. The crux of the matter for all zero-shot learning methods is to associate observed and non observed classes through some form of auxiliary information which encodes visually distinguishing properties of objects. Different flavors of zero-shot learning methods that we evaluate in this work are linear~\cite{FCSBDRM13,APHS13,ARWLS15,RT15} and nonlinear~\cite{XASNHS16,SGMN13} compatibility learning frameworks whereas an orthogonal direction is learning independent attribute~\cite{LNH13} classifiers and finally others~\cite{ZV15,CCGS16,NMBSSFCD14} propose a hybrid model between independent classifier learning and compatibility learning frameworks. 

We thoroughly evaluate the second aspect of zero-shot learning, by using multiple splits of several small to large-scale datasets~\cite{PH12, CaltechUCSDBirdsDataset, LNH13, FEHF09, ImageNet}. We emphasize that it is hard to obtain labeled training data for fine-grained classes of rare objects recognizing which requires expert opinion. Therefore, we argue that zero-shot learning methods should be evaluated mainly on least populated or rare classes. 

We propose a unified evaluation protocol to address the third aspect of zero-shot learning which is arguably the most important one. We emphasize the necessity of tuning hyperparameters of the methods on a validation class split that is disjoint from training classes as improving zero-shot learning performance via tuning parameters on test classes violates the zero-shot assumption. We argue that per-class averaged top-1 accuracy is an important evaluation metric when the dataset is not well balanced with respect to the number of images per class. We point out that extracting image features via a pre-trained deep neural network (DNN) on a large dataset that contains zero-shot test classes also violates the zero-shot learning idea as image feature extraction is a part of the training procedure. Moreover, we argue that demonstrating zero-shot performance on small-scale and coarse grained datasets, i.e. aPY~\cite{FEHF09} is not conclusive. We recommend to abstract away from the restricted nature of zero-shot evaluation and make the task more practical by including training classes in the search space, i.e. generalized zero-shot learning setting. Therefore, we argue that our work plays an important role in advancing the zero-shot learning field by analyzing the good and bad aspects of the zero-shot learning task as well as proposing ways to eliminate the ugly ones.

\section{Related Work}
\label{sec:related}

We review related work on zero-shot and generalized zero-shot learning, we present previous evaluations on the same task and describe the unique aspects of our work. 

\myparagraph{Zero-Shot Learning.} In zero-shot learning setting test and training class sets are disjoint~\cite{HEEY15,LNH13,LEB08,RSS11,YA10} which can be tackled by solving related sub-problems, \eg learning intermediate attribute classifiers~\cite{LNH13,RSS11,RSS10} and learning a mixture of seen class proportions~\cite{ZV15,ZV16,NMBSSFCD14,CCGS16}, or by a direct approach, \eg compatibility learning frameworks~\cite{APHS15,ARWLS15,FCSBDRM13,Hastie:Tibshirani:Friedman:2008,palatucci2009zero,RT15,SGMN13,XASNHS16,RT15,FS16, QLSH16, AMFS16,BHJ16, MGS14, FXKG15,KXFG15}. Among these methods, in our evaluation we choose to use DAP~\cite{LNH13} for being one of the most fundamental methods in zero-shot learning research; CONSE~\cite{NMBSSFCD14} for being one of the most widely used representatives of learning a mixture of class proportions;  SSE~\cite{ZV15} for  being a recent method with a public implementation; SJE~\cite{ARWLS15}, ALE~\cite{APHS15}, DEVISE~\cite{FCSBDRM13} for being recent compatibility learning methods with similar loss functions; ESZSL~\cite{RT15} for adding a regularization term to unregularized compatibility learning methods; \cite{XASNHS16} and CMT~\cite{SGMN13} proposing non-linear extensions to bilinear compatibility learning framework and finally SYNC~\cite{CCGS16} for reporting the state-of-the-art on several benchmark datasets.

\myparagraph{Generalized Zero-shot Learning.} This setting~\cite{SRSB13} generalizes the zero-shot learning task to the case with both seen and unseen classes at test time. \cite{JSB14} argues that although ImageNet classification challenge performance has reached beyond human performance, we do not observe similar behavior of the methods that compete at the detection challenge which involves rejecting unknown objects while detecting the position and label of a known object. \cite{FCSBDRM13} uses label embeddings to operate on the generalized zero-shot learning setting whereas \cite{ZSYXLC16} proposes to learn latent representations for images and classes through coupled linear regression of factorized joint embeddings. On the other hand, \cite{BB16} introduces a new model layer to the deep net which estimates the probability of an input being from an unknown class and \cite{SGMN13} proposes a novelty detection mechanism. We evaluate \cite{SGMN13} and \cite{FCSBDRM13} for being the most widely used.

\myparagraph{Previous Evaluations of Zero-Shot Learning.} In the literature some zero-shot vs generalized zero-shot learning evaluation works exist~\cite{RSS11, CCGS16b}. Among these, \cite{RSS11} proposes a model to learn the similarity between images and semantic embeddings on the ImageNet 1K by using 800 classes for training and 200 for test. \cite{CCGS16b} provides a comparison between five methods evaluated on three datasets including ImageNet with three standard splits and proposes a metric to evaluate generalized zero-shot learning performance.

\myparagraph{Our work.} We evaluate ten zero-shot learning methods on five datasets with several splits both for zero-shot and generalized zero-shot learning settings, provide statistical significancy and robustness tests, and present other valuable insights that emerge from our benchmark. In this sense, ours is a more extensive evaluation compared to prior work.

\section{Evaluated Methods}
\label{sec:method}
We start by formalizing the zero-shot learning task and then we describe the zero-shot learning methods that we evaluate in this work. Given a training set $\mathcal{S} = \{(x_n, y_n), n=1...N\}$, with $y_n \in \mathcal{Y}^{tr}$ belonging to training classes, the task is to learn $f: \mathcal{X} \rightarrow \mathcal{Y}$ by minimizing the regularized empirical risk:
\begin{equation}\label{eq:compatibility}
\frac{1}{N}\sum_{n=1}^N L(y_n, f(x_n; W)) + \Omega(W)
\end{equation}
with $L(.)$ being the loss function and $\Omega(.)$ being the regularization term.
Here, the mapping $f: \mathcal{X} \rightarrow \mathcal{Y}$ from input to output embeddings is defined as:
\begin{equation}
f(x;W) = \argmax_{y\in\mathcal{Y}} F(x,y;W) 
\end{equation}
At test time, in zero-shot learning setting, the aim is to assign a test image to an unseen class label, i.e. $\mathcal{Y}^{ts}\subset \mathcal{Y}$ and in generalized zero-shot learning setting, the test image can be assigned either to seen or unseen classes, i.e. $\mathcal{Y}^{tr+ts}\subset \mathcal{Y}$ with the highest compatibility score. 

\subsection{Learning Linear Compatibility}
\label{subsec:compatibilty}
Attribute Label Embedding (ALE)~\cite{APHS15}, Deep Visual Semantic Embedding (DEVISE)~\cite{FCSBDRM13} and Structured Joint Embedding (SJE)~\cite{ARWLS15} use bi-linear compatibility function to associate visual and auxiliary information:
\begin{equation}\label{eq:comp}
F(x,y;W) = \theta(x)^T W \phi(y)
\end{equation}
where $\theta(x)$ and $\phi(y)$, i.e. image and class embeddings, both of which are given. $F(.)$ is parameterized by the mapping $W$, to be learned. Embarassingly Simple Zero Shot Learning (ESZSL)~\cite{RT15} adds a regularization term to this objective. In the following, we provide a unified formulation of these four zero-shot learning methods.

\myparagraph{DEVISE~\cite{FCSBDRM13}} uses pairwise ranking objective that is inspired from unregularized ranking SVM~\cite{J02}:
\begin{equation}\label{eq:rank}
\sum_{y\in \mathcal{Y}^{tr}}[\Delta(y_n,y) + F(x_n,y;W) - F(x_n,y_n;W)]_+
\end{equation}

\myparagraph{ALE~\cite{APHS15}} uses weighted approximate ranking objective~\cite{UBG09}:
\begin{equation}
\sum_{y\in \mathcal{Y}^{tr}} \frac{l_{r_{\Delta(x_n,y_n)}}}{r_{\Delta(x_n,y_n)}} [\Delta(y_n,y) + F(x_n,y;W) - F(x_n,y_n;W)]_+
\end{equation}
where $l_k= \sum_{i=1}^{k}\alpha_i$ and $r_{\Delta(x_n,y_n)}$ is defined as:
\begin{equation}
\sum_{y \in \mathcal{Y}^{tr}} \mathds{1}(F(x_n,y;W) + \Delta(y_n,y) \geq F(x_n,y_n;W))
\end{equation} 
Following the heuristic in \cite{WBU10}, \cite{APHS15} selects $\alpha_i = 1/i$ which puts high emphasis on the top of the rank list. 

\myparagraph{SJE~\cite{ARWLS15}} gives full weight to the top of the ranked list and is inspired from the structured SVM~\cite{TJH05}:
\begin{equation}
[\max_{y\in \mathcal{Y}^{tr}}(\Delta(y_n,y) + F(x_n,y;W)) - F(x_n,y_n;W)]_+
\end{equation}

\myparagraph{ESZSL~\cite{RT15}} adds the following regularization term to the unregularized risk minimization formulation:
\begin{equation}
\gamma\|W\phi(y)\|^2_{Fro} + \lambda\|\theta(x)^TW\|^2_{Fro} + \beta\|W\|^2_{Fro}
\end{equation}
where $\gamma, \lambda, \beta$ are parameters of this regularizer.

\subsection{Learning Nonlinear Compatibility}
Latent Embeddings (LATEM)~\cite{XASNHS16} and Cross Modal Transfer (CMT)~\cite{SGMN13} encode an additional non-linearity in compatibility learning framework.

\myparagraph{LATEM~\cite{XASNHS16}} constructs a piece-wise linear compatibility:
\begin{equation}
F(x,y;W_i) = \max_{1\leq i\leq K}\theta(x)^T W_i \phi(y)
\end{equation}
where every $W_i$ models a different visual characteristic of the data and the selection of which matrix to use to do the mapping is a latent variable. LATEM uses the ranking loss formulated in~\autoref{eq:rank}.

\myparagraph{CMT~\cite{SGMN13}} first maps images into a semantic space of words, i.e. class names, where a neural network with $\tanh$ nonlinearity learns the mapping:
\begin{equation}
\sum_{y\in \mathcal{Y}^{tr}} \sum_{x\in\mathcal{X}_y} \|\phi(y) - W_1 \tanh(W_2. \theta(x)\|
\end{equation}
where $(W_1,W_2)$ are weights of the two layer neural network. This is followed by a novelty detection mechanism that assigns images to unseen or seen classes. The novelty is detected either via thresholds learned using the embedded images of the seen classes or the outlier probabilities are obtained in an unsupervised way.

\subsection{Learning Intermediate Attribute Classifiers}
Although Direct Attribute Prediction (DAP)~\cite{LNH13} has been shown to perform poorly compared to compatibility learning frameworks~\cite{APHS15}, we include it to our evaluation for being historically one of the most widely used methods in the literature. 

\myparagraph{DAP~\cite{LNH13}} learns probabilistic attribute classifiers and makes a class prediction by combining scores of the learned attribute classifiers. A novel image is assigned to one of the unknown classes using:
\begin{equation}
f(x) = \argmax_{c} \prod_{m=1}^M \frac{p(a_m^c|x)}{p(a_m^c)}.
\end{equation}
with $M$ being the total number of attributes. We train a one-vs-rest SVM with log loss that gives probability scores of attributes with respect to training classes.

\subsection{Hybrid Models}

Semantic Similarity Embedding (SSE)~\cite{ZV15}, Convex Combination of Semantic Embeddings (CONSE)~\cite{NMBSSFCD14} and Synthesized Classifiers (SYNC)~\cite{CCGS16} express images and semantic class embeddings as a mixture of seen class proportions, hence we group them as hybrid models.

\myparagraph{SSE~\cite{ZV15}} leverages similar class relationships both in image and semantic embedding space. An image is labeled with:
\begin{equation}
\argmax_{u\in \mathcal{U}} \pi(\theta(x))^T \psi(\phi(y_u))
\end{equation}
where $\pi, \psi$ are mappings of class and image embeddings into a common space. Specifically, $\psi$ is learned by sparse coding and $\pi$ is by class dependent transformation.

\myparagraph{CONSE~\cite{NMBSSFCD14}} learns the probability of a training image belonging to a training class:
\begin{equation}
f(x,t) = \argmax_{y\in\mathcal{Y}^{tr}}p_{tr}(y|x)
\end{equation}
where $y$ denotes the most likely training label ($t$=1) for image $x$. Combination of semantic embeddings ($s$) is used to assign an unknown image to an unseen class:
\begin{equation}
\frac{1}{Z} \sum_{i=1}^T p_{tr}(f(x,t)|x) . s(f(x,t))
\end{equation}
where $Z = \sum_{i=1}^T p_{tr}(f(x,t)|x)$, $f(x,t)$ denotes the t$^{th}$ most likely label for image $x$ and $T$ controls the maximum number of semantic embedding vectors.

\myparagraph{SYNC~\cite{CCGS16}} learns a mapping between the semantic class embedding space and a model space. In the model space, training classes and a set of phantom classes form a weighted bipartite graph. The objective is to minimize distortion error:
\begin{equation}
\min_{w_c,v_r} \|w_c - \sum_{r=1}^R s_{cr} v_r \|_2^2.
\end{equation}
Semantic and model spaces are aligned by embedding real ($w_c$) and phantom classes ($v_r$) in the weighted graph ($s_{cr}$). 


{
\setlength{\tabcolsep}{4.5pt}
\renewcommand{\arraystretch}{1.2}
\begin{table*}[t]
\vspace{-3mm}
 \begin{center}
  \begin{tabular}{ l c c c  c c c | c c c c c  c c c c|}
    & & & & \multicolumn{3}{c}{\multirow{3}{*}{\textbf{Number of Classes}}}& \multicolumn{9}{c}{\textbf{Number of Images}} \\
    \cline{8-16}
    & & & &  & & & & \multicolumn{4}{c}{\textbf{At Training Time}} & \multicolumn{4}{c|}{\textbf{At Evaluation Time}} \\
    & & & & & & & & \multicolumn{2}{c}{\textbf{SS}} & \multicolumn{2}{c}{\textbf{PS}} & \multicolumn{2}{c}{\textbf{SS}} & \multicolumn{2}{c|}{\textbf{PS}} \\
    \textbf{Dataset} & \textbf{Size} & \textbf{Detail} & \textbf{Att} & $\mathcal{Y}$ & $\mathcal{Y}^{tr}$ & $\mathcal{Y}^{ts}$ & \textbf{Total} & $\mathcal{Y}^{tr}$ & $\mathcal{Y}^{ts}$ & $\mathcal{Y}^{tr}$ & $\mathcal{Y}^{ts}$ & $\mathcal{Y}^{tr}$ & $\mathcal{Y}^{ts}$ & $\mathcal{Y}^{tr}$ & $\mathcal{Y}^{ts}$ \\     \hline
    SUN~\cite{PH12} & medium & fine & 102 & 717 & $580$ + $65$ & 72 & 14K & $12900$ & $0$ & $10320$ & $0$ &$0$ & $1440$ & $2580$ & $1440$ \\
    CUB~\cite{CaltechUCSDBirdsDataset} & medium & fine  & 312 & 200 & 100 + 50 & 50 & 11K & $8855$ & $0$ & $7057$ & $0$ &$0$ & $2933$ & $1764$ & $2967$ \\
    AWA~\cite{LNH13} & medium & coarse  & 85 & 50 & 27 + 13 & 10 & 30K & $24295$ & $0$ & $19832$ & $0$ &$0$ & $6180$ & $4958$ & $5685$ \\
    aPY~\cite{FEHF09} & small & coarse & 64 & 32 & $15$ + $5$ & 12 & 15K & $12695$ & $0$ & $5932$ & $0$ &$0$ & $2644$ & $1483$ & $7924$ \\ 
    \hline %
  \end{tabular}
 \end{center}
 \vspace{-5mm}
\caption{Statistics for attribute datasets: SUN~\cite{PH12}, CUB~\cite{CaltechUCSDBirdsDataset}, AWA~\cite{LNH13}, aPY~\cite{FEHF09} in terms of size of the datasets, fine-grained or coarse-grained, number of attributes, number of classes in training + validation ($\mathcal{Y}^{tr}$) and test classes ($\mathcal{Y}^{ts}$), number of images at training and test time for standard split (SS) and our proposed splits (PS).}
\vspace{-2mm}
\label{tab:datasets}
\end{table*}
}

\section{Datasets and Evaluation Protocol}
\label{sec:datasets}
In this section, we provide several components of previously used and our proposed zero-shot and generalized zero-shot learning evaluation protocols, e.g. datasets, image and class encodings and the evaluation protocol.

\subsection{Dataset Statistics}
Among the most widely used datasets for zero-shot learning, we select two coarse-grained, one small and one medium-scale, and two fine-grained, both medium-scale, datasets with attributes and one large-scale dataset without. Here, we consider  between $10K$ and $1M$ images, and, between $100$ and $1K$ classes as medium-scale. 

\myparagraph{Attribute Datasets.} Statistics of the attribute datasets are presented in~\autoref{tab:datasets}.
Attribute Pascal and Yahoo (aPY)~\cite{FEHF09} is a small-scale coarse-grained dataset with $64$ attributes. Among the total number of $32$ classes, $20$ Pascal classes are used for training (we randomly select $5$ for validation) and $12$ Yahoo classes are used for testing.
Animals with Attributes (AWA)~\cite{LNH13} is a coarse-grained dataset that is medium-scale in terms of the number of images, i.e. $30,475$ and small-scale in terms of number of classes, i.e. $50$. \cite{LNH13} introduces a standard zero-shot split with $40$ classes for training (we randomly select $13$ for validation) and $10$ for testing. AWA has $85$ attributes. 
Caltech-UCSD-Birds 200-2011 (CUB)~\cite{CaltechUCSDBirdsDataset} is a fine-grained and medium scale dataset with respect to both number of images and number of classes, i.e. $11,788$ images from $200$ different types of birds annotated with $312$ attributes. \cite{APHS15} introduces the first zero-shot split of CUB with $150$ training ($50$ validation classes) and $50$ test classes. 
SUN~\cite{PH12} is a fine-grained and medium-scale dataset with respect to both number of images and number of classes, i.e. SUN contains $14340$ images coming from $717$ types of scenes annotated with $102$ attributes. Following~\cite{LNH13} we use $645$ classes of SUN for training (we randomly select $65$ for val) and $72$ for testing. 

\myparagraph{Large-Scale ImageNet.} We also evaluate the performance of methods on the large scale ImageNet~\cite{ImageNet}. Among the total of 21K classes, 1K classes are used for training (we use $200$ classes for validation) and the test split is either all the remaining 21K classes or a subset of it, e.g. we determine these subsets based on the hierarchical distance between classes and the population of classes.

\subsection{Proposed Evaluation Protocol} 
We present our proposed unified protocol for image and class embeddings, dataset splits and evaluation criteria.

\myparagraph{Image and Class Embedding.}
We extract image features from the entire image for SUN, CUB, AWA and ImageNet, with no image pre-processing. For aPY, as proposed in~\cite{FEHF09}, we extract image features from bounding boxes. Our image embeddings are $2048$-dim top-layer pooling units of the $101$-layered ResNet~\cite{HZRS15} as we found that it performs better than $1,024$-dim top-layer pooling units of GoogleNet~\cite{SLJSRAEVR15}. ResNet is pre-trained on ImageNet 1K and not fine-tuned. In addition to ResNet features, we evaluate methods with their published image features. As class embeddings, for aPY, AWA, CUB and SUN, we use per-class attributes. For ImageNet we use Word2Vec~\cite{MSCCD13} provided by~\cite{CCGS16} as it does not contain attribute annotation for all the classes.

\myparagraph{Dataset Splits.}
Zero-shot learning assumes disjoint training and test classes with the presence of all the images of training classes and the absence of any image from test classes during training. On the other hand, as deep neural network (DNN) training for image feature extraction is actually a part of model training, the dataset used to train DNNs, e.g. ImageNet, should not include any of the test classes. However, we notice from the standard splits (SS) of aPY and AWA datasets that 7 aPY test classes out of 12 (monkey, wolf, zebra, mug, building, bag, carriage), 6 AWA test classes out of 10 (chimpanzee, giant panda, leopard, persian cat, pig, hippopotamus), are among the 1K classes of ImageNet, i.e. are used to pre-train ResNet. On the other hand, the mostly widely used splits, i.e. we term them as standard splits (SS), for SUN from~\cite{LNH13} and CUB from~\cite{APHS13} shows us that 1 CUB test class out of 50 (Indigo Bunting), and 6 SUN test classes out of 72 (restaurant, supermarket, planetarium, tent, market, bridge), are also among the 1K classes of ImageNet. We noticed that the accuracy for all methods on those overlapping test classes are higher than others. Therefore, we propose new dataset splits, i.e. proposed splits (PS), insuring that none of the test classes appear in ImageNet 1K, i.e. used to train the ResNet model. We present the differences between the standard splits (SS) and the proposed splits (PS) in~\autoref{tab:datasets}. While in SS and PS no image from test classes is present at training time, at test time SS does not include any images from training classes however our PS does. We designed the PS this way as evaluating accuracy on both training and test classes is crucial to show the generalization of methods. \modif{Note that we introduce Proposed Split Version 2.0\footnote{\url{http://www.mpi-inf.mpg.de/zsl-benchmark}}.}

ImageNet with thousands of classes provides possibilities of constructing several zero-shot evaluation splits. Following~\cite{CCGS16}, our first two standard splits consider all the classes that are 2-hops and 3-hops away from the original 1K classes according to the ImageNet label hierarchy, corresponding to $1509$ and $7678$ classes. This split measures the generalization ability of the models with respect to the hierarchical and semantic similarity between classes. Our proposed split considers 500, 1K and 5K most populated classes among the remaining 21K classes of ImageNet with $\approx 1756$, $\approx 1624$ and $\approx 1335$ images per class on average. Similarly, we consider 500, 1K and 5K least-populated classes in ImageNet which correspond to most fine-grained subsets of ImageNet with $\approx 1$, $\approx 3$ and $\approx 51$ images per class on average. Our final split considers all the remaining $\approx 20$K classes of ImageNet with at least $1$ image per-class, $\approx 631$ images per class on average.

\myparagraph{Evaluation Criteria.}
Single label image classification accuracy has been measured with Top-1 accuracy, i.e. the prediction is accurate when the predicted class is the correct one. If the accuracy is averaged for all images, high performance on densely populated classes is encouraged. However, we are interested in having high performance also on sparsely populated classes. Therefore, we average the correct predictions independently for each class before dividing their cumulative sum w.r.t the number of classes, i.e. we measure average per-class top-1 accuracy. 

In generalized zero-shot learning setting, the search space at evaluation time is not restricted to only test classes, but includes also the training classes, hence this setting is more practical. As with our proposed split at test time we have access to some images from training classes, after having computed the average per-class top-1 accuracy on training and test classes, we compute the harmonic mean of training and test accuracies:
\begin{equation}
H = 2* (acc_{\mathcal{Y}^{tr}}*acc_{\mathcal{Y}^{ts}})/(acc_{\mathcal{Y}^{tr}}+acc_{\mathcal{Y}^{ts}})
\end{equation}
where $acc_{\mathcal{Y}^{tr}}$ and $acc_{\mathcal{Y}^{ts}}$ represent the accuracy of images from seen ($\mathcal{Y}^{tr}$), and images from unseen ($\mathcal{Y}^{ts}$) classes respectively. We choose harmonic mean as our evaluation criteria and not arithmetic mean because in arithmetic mean if the seen class accuracy is much higher, it effects the overall results significantly. Instead, our aim is high accuracy on both seen and unseen classes.

{
\renewcommand{\arraystretch}{1.1}
\begin{table}[t]
 \centering
   \begin{tabular}{l c c c c}
      & \multicolumn{2}{c }{\textbf{SUN}} & \multicolumn{2}{c }{\textbf{AWA}} \\
      \textbf{Model} & \textbf{R} & \textbf{O}   & \textbf{R} & \textbf{O} \\     
     \hline
     DAP~\cite{LNH13} & $22.1$ & $22.2$ & $41.4$ & $41.4$ \\
     SSE~\cite{ZV15} & $83.0$ & $82.5$ & $64.9$ & $76.3$ \\
     LATEM~\cite{XASNHS16} & -- & -- & $71.2$ & $71.9$ \\
     SJE~\cite{ARWLS15} & -- & -- & $67.2$ & $66.7$ \\
     ESZSL~\cite{RT15} & $64.3$ & $65.8$ & $48.0$ & $49.3$ \\
     SYNC~\cite{CCGS16} & $62.8$ & $62.8$& $69.7$ & $69.7$ \\ 
     \hline
   \end{tabular} 
   \vspace{-1mm}
\caption{Reproducing zero-shot results: O = Original results published in the paper, R = Reproduced using provided image features and code. We measure top-1 accuracy in \%.}
\vspace{-3mm}
\label{tab:reproduce}
\end{table}
}

\section{Experiments}
\label{sec:exp}
We first provide zero-shot learning results on attribute datasets SUN, CUB, AWA and aPY and then on the large-scale ImageNet dataset. Finally, we present results for the generalized zero-shot learning setting.

\subsection{Zero-Shot Learning Results}
On attribute datasets, i.e. SUN, CUB, AWA and aPY, we first reproduce the results of each method using their evaluation protocol, then provide a unified evaluation protocol using the same train/val/test class splits, followed by our proposed train/val/test class splits. We also evaluate the robustness of the methods to parameter tuning and visualize the ranking of different methods. Finally, we evaluate the methods on the large-scale ImageNet dataset.

\myparagraph{Reproducing Results.} For sanity-check, we re-evaluate methods \cite{LNH13,ZV15,XASNHS16,ARWLS15,RT15,CCGS16}~\footnote{\cite{SGMN13} has public code available, but is not evaluated on SUN or AWA.} using provided features and code. We chose SUN and AWA as two representative of fine-grained and non-fine-grained datasets having been widely used in the literature. We observe from the results in~\autoref{tab:reproduce} that our reproduced results and the reported results of DAP and SYNC are identical to the reported number in their original publications. For LATEM, we obtain slightly different results which can be explained by the non-convexity and thus the sensibility to initialization. Similarly for SJE random sampling in SGD might lead to slightly different results. ESZSL has some variance because its algorithm randomly picks a validation set during each run, which leads to different hyperparameters. Notable observations on SSE~\cite{ZV15} results are as follows. The published code has hard-coded hyperparameters operational on aPY, i.e. number of iterations, number of data points to train SVM, and one regularizer parameter $\gamma$ which lead to inferior results than the ones reported here, therefore we set these parameters on validation sets. On SUN, SSE uses $10$ classes (instead of $72$) and our results with validated parameters got an improvement of $0.5\%$ that may be due to random sampling of training images. On AWA, our reproduced result being $64.9\%$ is significantly lower than the reported result ($76.3\%$). However, we could not reach the reported result even by tuning parameters on the test set, i.e. we obtain $73.8\%$ in this case.

{
\setlength{\tabcolsep}{2pt}
\renewcommand{\arraystretch}{1.2}
\begin{table}[t]
 \centering
 \resizebox{\linewidth}{!}{%
   \begin{tabular}{l c c |c c |c c |c c }
      & \multicolumn{2}{c }{\textbf{SUN}} & \multicolumn{2}{c }{\textbf{CUB}} & \multicolumn{2}{c }{\textbf{AWA}} & \multicolumn{2}{c }{\textbf{aPY}}   \\
      \textbf{Method} & \textbf{SS} & \textbf{PS} & \textbf{SS} & \textbf{PS} & \textbf{SS} & \textbf{PS} & \textbf{SS} & \textbf{PS} \\     
     \hline
     DAP~\cite{LNH13}  & $38.9$ & $39.9$ & $37.5$ & $40.0$ & $57.1$ & $44.1$ & $35.2$ & $33.8$ \\
     CONSE~\cite{NMBSSFCD14} & $44.2$ & $38.0$ & $36.7$ & $33.6$ & $63.6$ & $46.3$ & $25.9$ & $26.4$\\
     CMT~\cite{SGMN13} & $41.9$ & $40.1$ & $37.3$ & $34.6$ & $58.9$ & $39.5$ & $26.9$ & $28.0$\\
     SSE~\cite{ZV15} & $54.5$ & $51.5$ & $43.7$ & $43.9$ & $68.8$ & $60.1$ & $31.1$ & $35.0$    \\
     LATEM~\cite{XASNHS16} & $56.9$ & $55.3$ & $49.4$ & $49.6$ & $74.8$ & $55.1$ & $34.5$ & $36.8$     \\
     ALE~\cite{APHS15} & $59.1$ & $\mathbf{58.1}$ & $53.2$ & $54.9$ & $\mathbf{78.6}$ & $59.9$ & $30.9$ & $\mathbf{39.7}$    \\
     DEVISE~\cite{FCSBDRM13}  & $57.5$ & $56.5$ & $53.2$ & $52.0$ & $72.9$& $54.2$ & $35.4$ & $37.0$    \\
     SJE~\cite{ARWLS15} & $57.1$ & $52.7$ & $\mathbf{55.3}$ & $53.9$ & $76.7$ & $\mathbf{65.6}$ & $32.0$ & $31.7$  \\
     ESZSL~\cite{RT15}  & $57.3$ & $54.5$ & $55.1$ & $51.9$ & $74.7$ & $58.2$ & $34.4$ & $38.3$    \\
     SYNC~\cite{CCGS16} & $\mathbf{59.1}$ & $56.2$ & $54.1$ & $\mathbf{56.0}$ & $72.2$ & $51.8$ & $\mathbf{39.7}$ & $23.9$   \\ 
     \hline
   \end{tabular} 
   }
   \vspace{-3mm}
\caption{Zero-shot on SS = Standard Split, PS = Proposed Split Version 2.0 using ResNet features (top-1 accuracy).}
\vspace{-3mm}
\label{tab:zeroshot}
\end{table}
}
\begin{figure}[t]
	\centering
	\includegraphics[width=0.48\columnwidth, trim=0 60 70 0,clip]{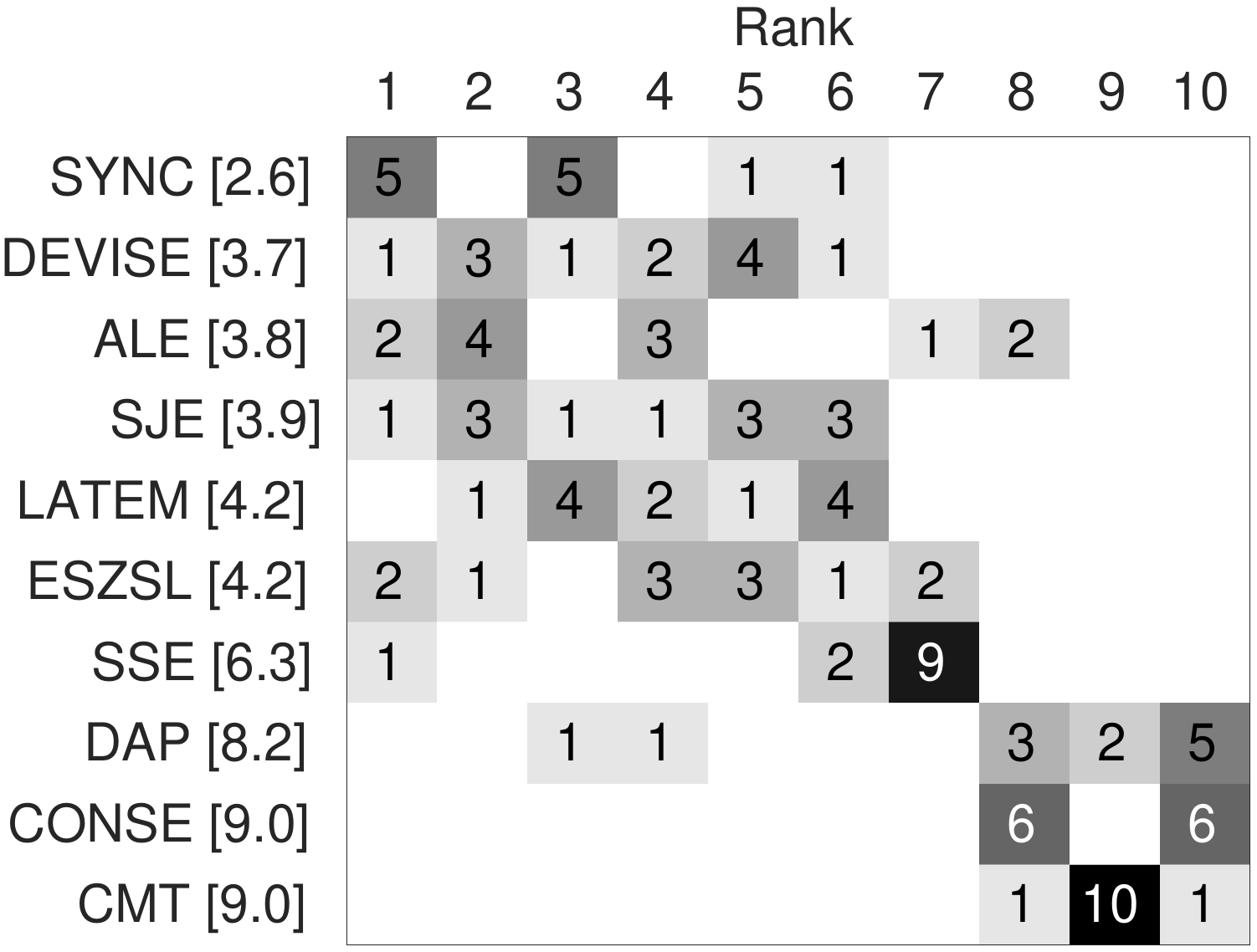}
    \includegraphics[width=0.48\columnwidth, trim=0 60 70 0,clip]{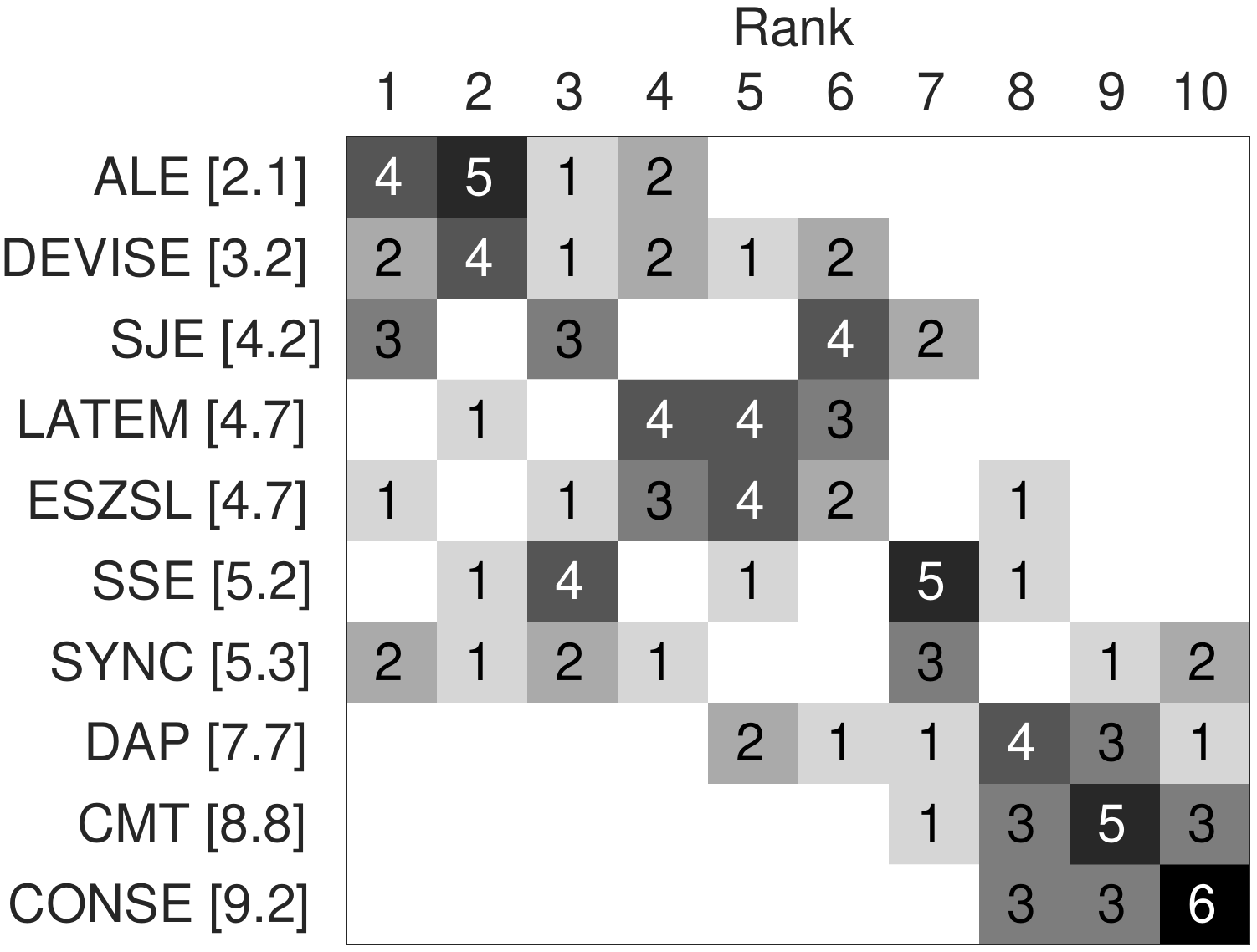}
    \vspace{-2mm}
    \caption{Ranking 10 models by setting parameters on three validation splits on the standard (SS, left) and proposed (PS Version 2.0, right) setting. Element $(i, j)$ indicates number of times model $i$ ranks at $j$th over all $4\times3$ observations. Models are ordered by their mean rank ( in brackets).} \vspace{-3mm}
	\label{fig:rank_matrix}
\end{figure}
\begin{figure*}[t]
	\centering 
    \includegraphics[width=0.23\linewidth, trim=20 5 70 10,clip]{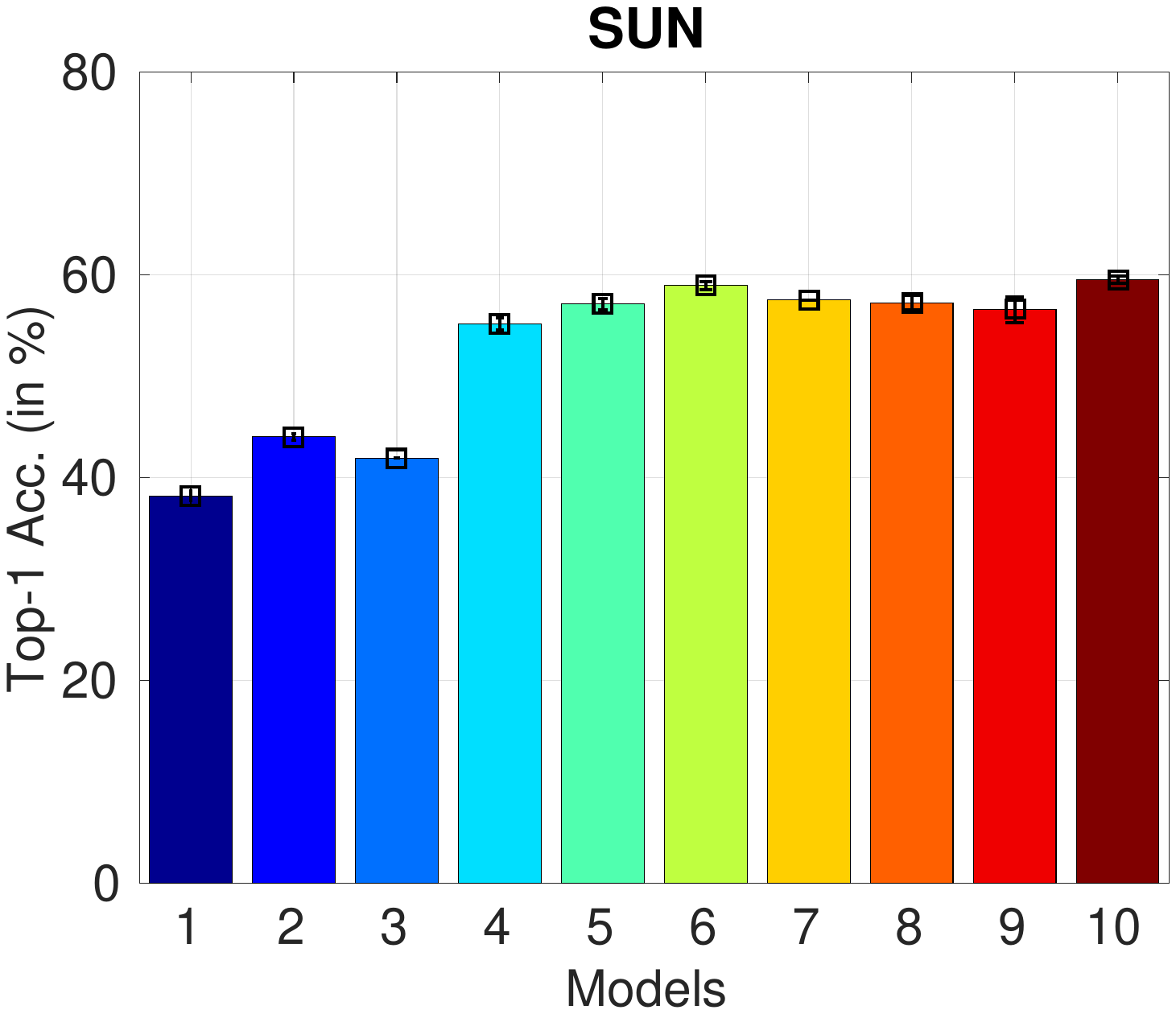}
    \includegraphics[width=0.23\linewidth, trim=20 5 70 10,clip]{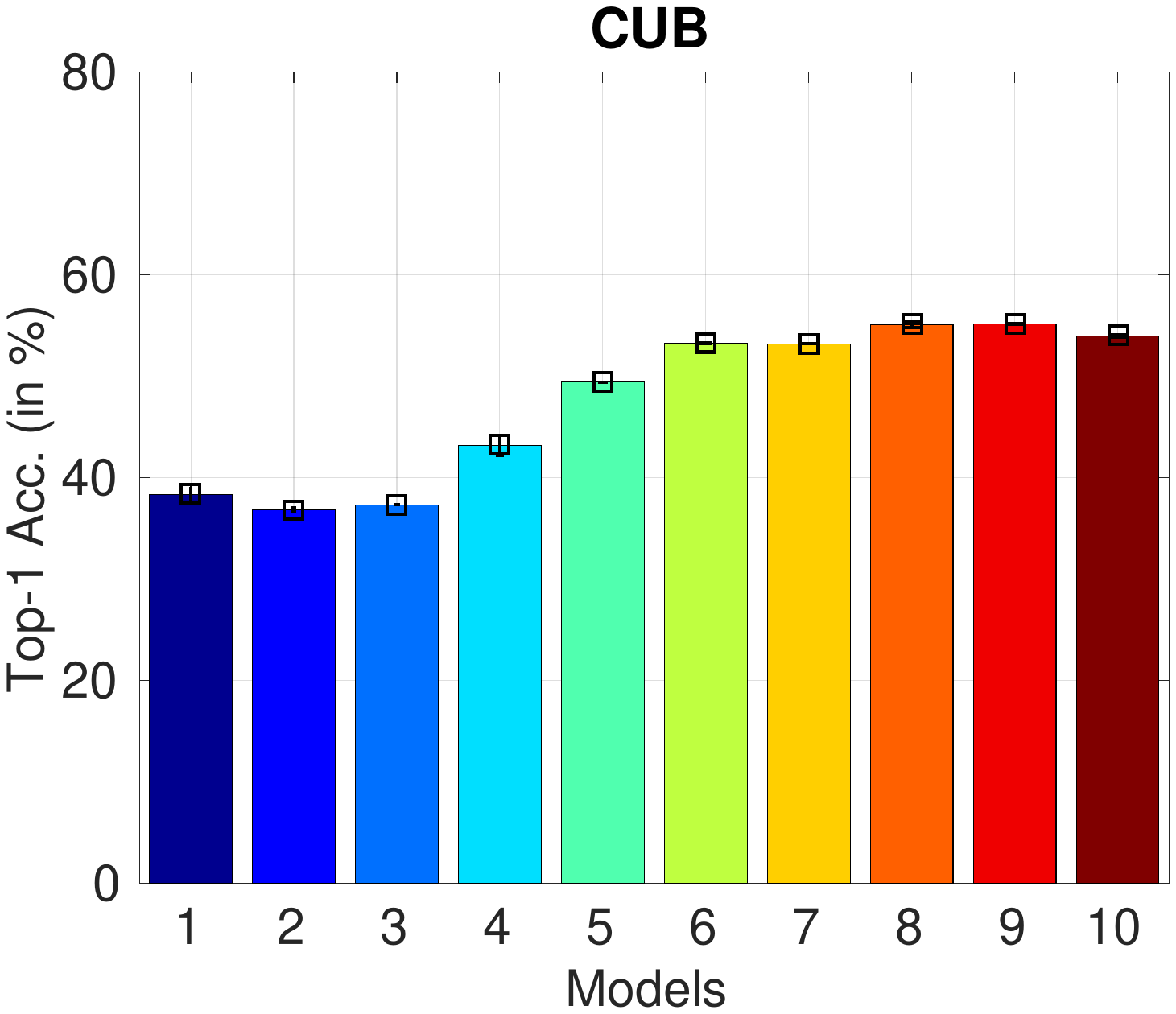}
    \includegraphics[width=0.23\linewidth, trim=20 5 70 10,clip]{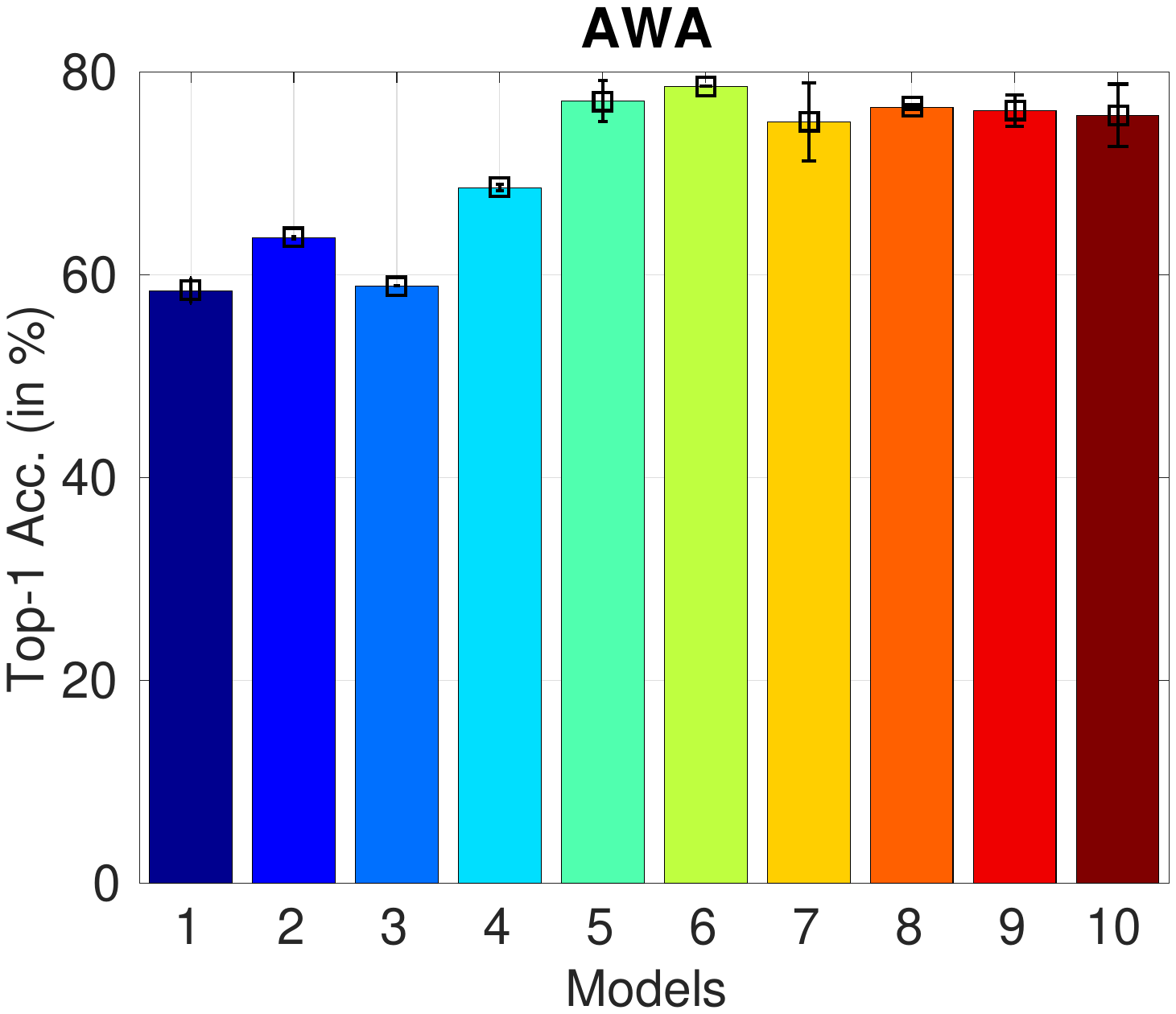}
    \includegraphics[width=0.23\linewidth, trim=20 5 70 10,clip]{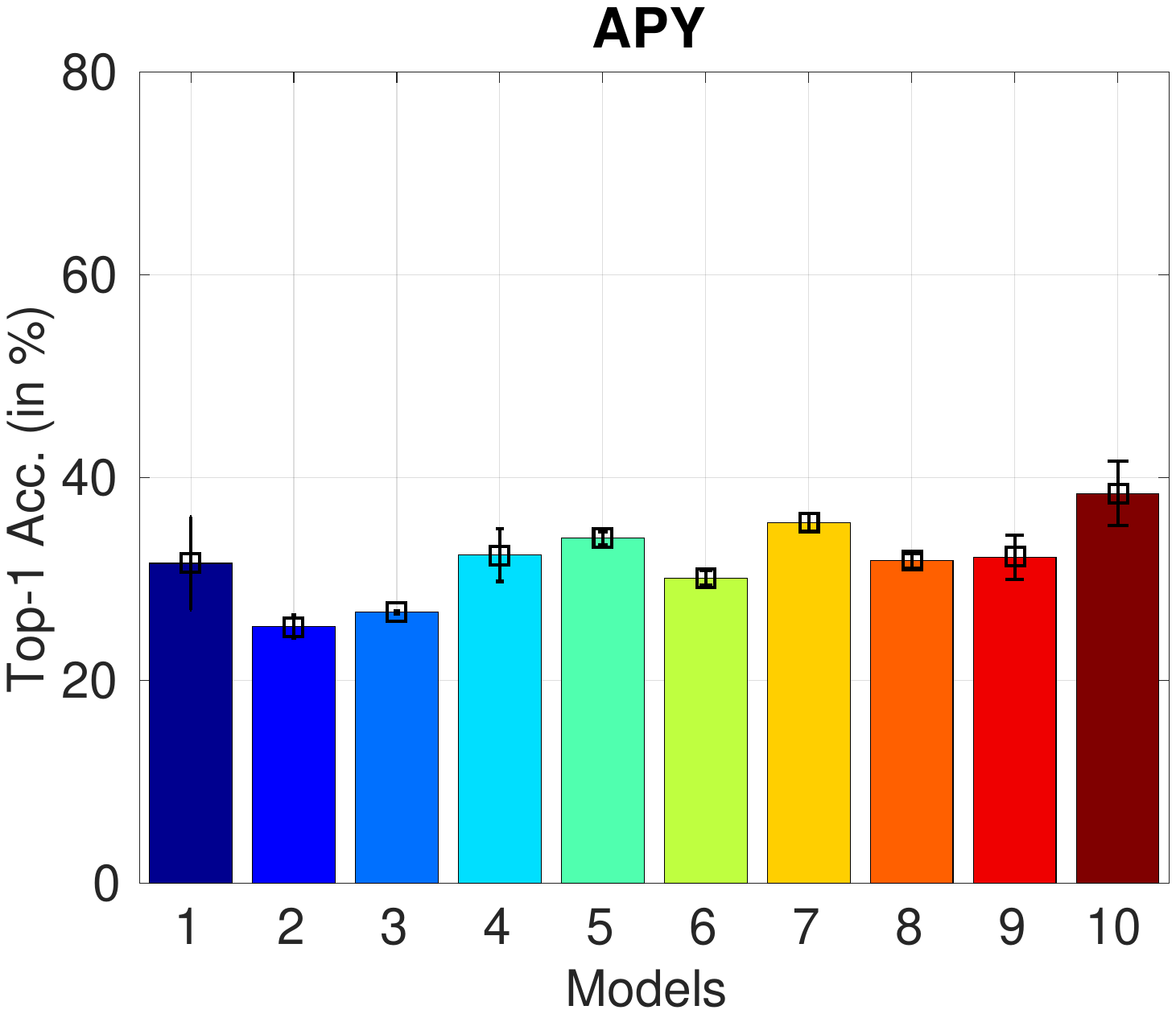}
    \includegraphics[width=0.05\linewidth, trim=570 5 55 10,clip]{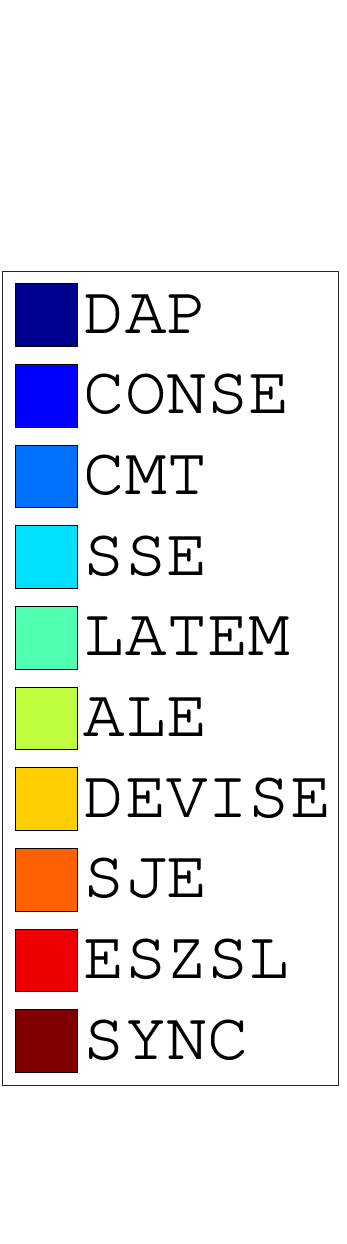}\\
    \includegraphics[width=0.23\linewidth, trim=20 5 70 10,clip]{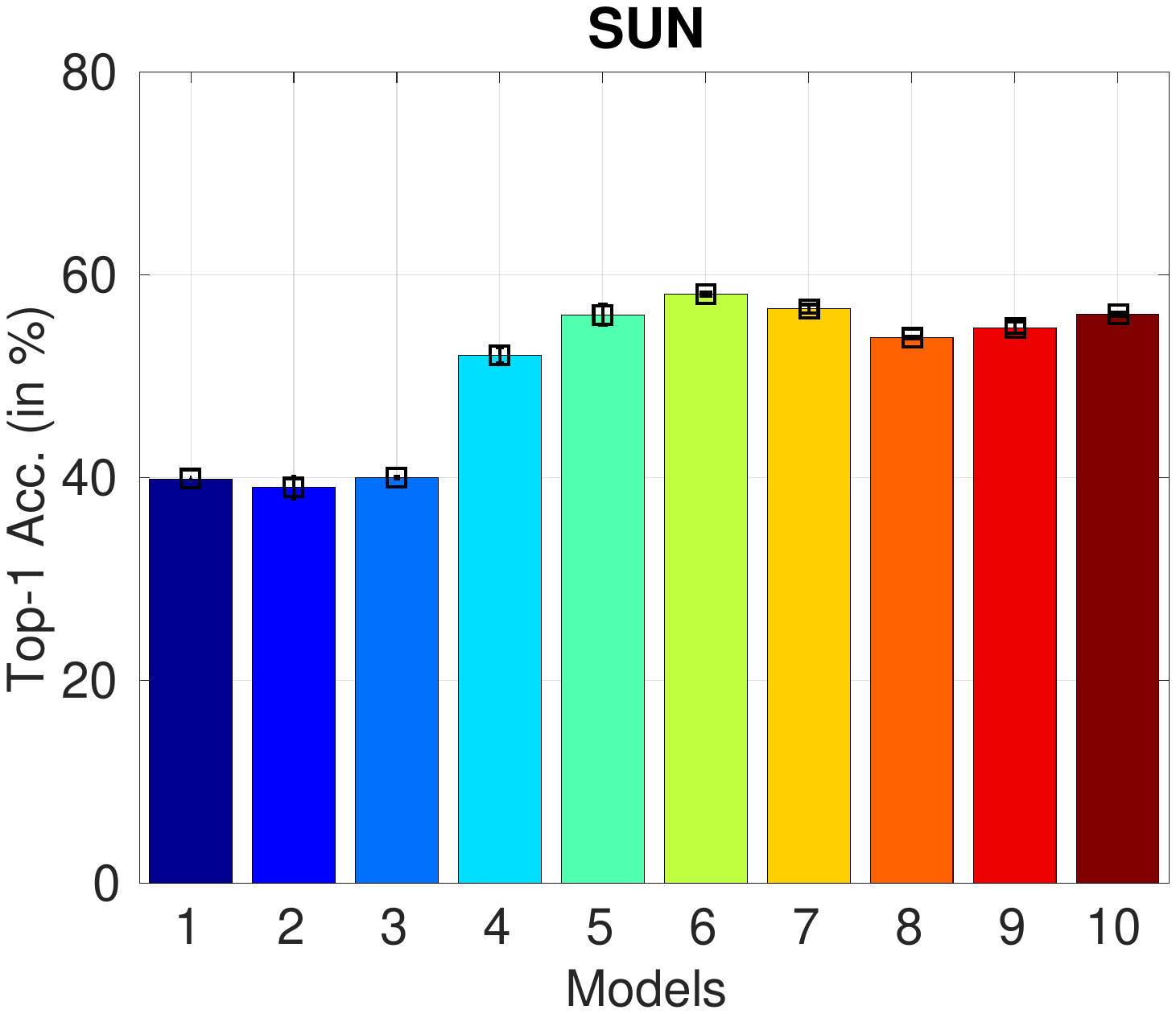}
	\includegraphics[width=0.23\linewidth, trim=20 5 70 10,clip]{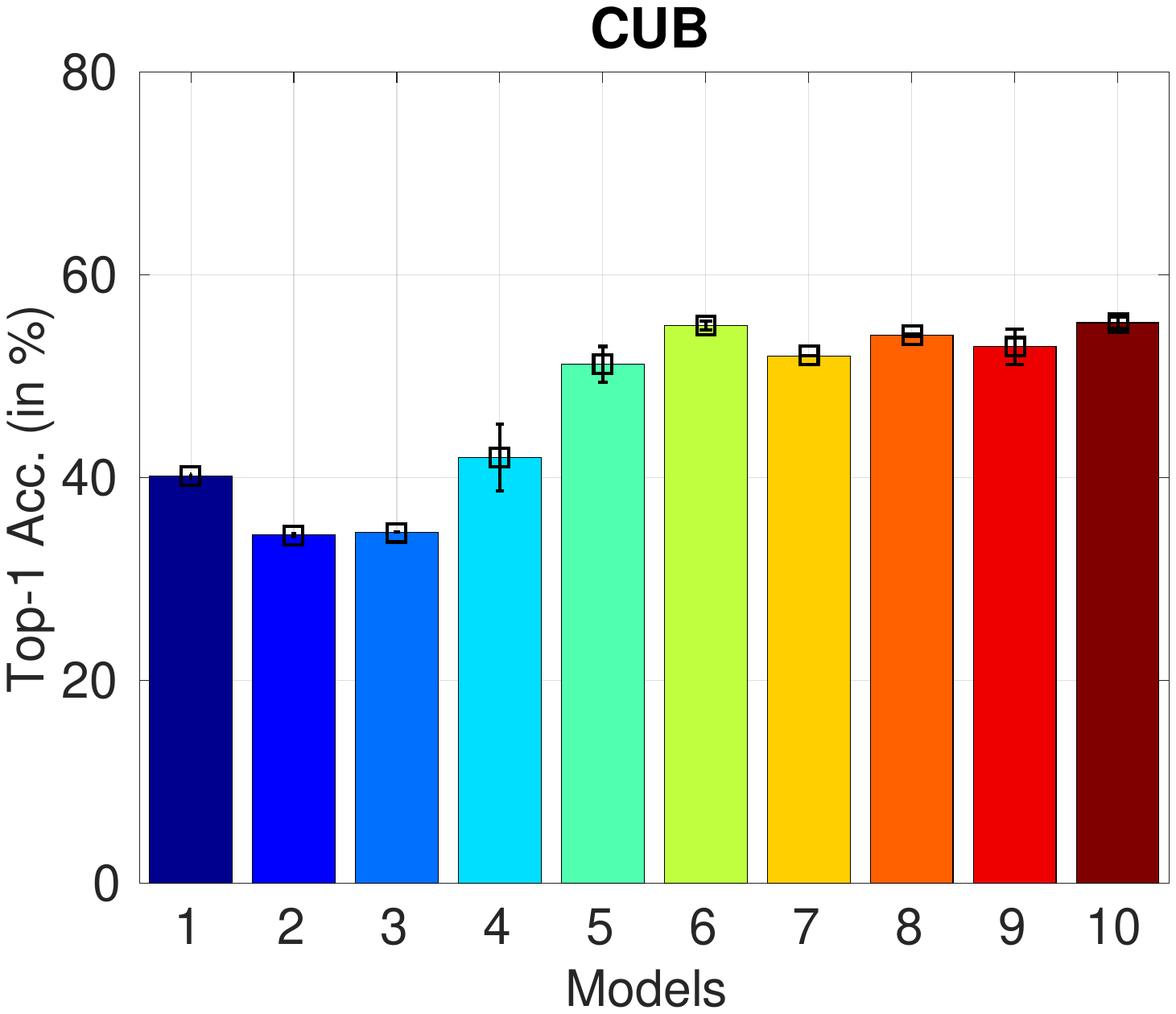}   
    \includegraphics[width=0.23\linewidth, trim=20 5 70 10,clip]{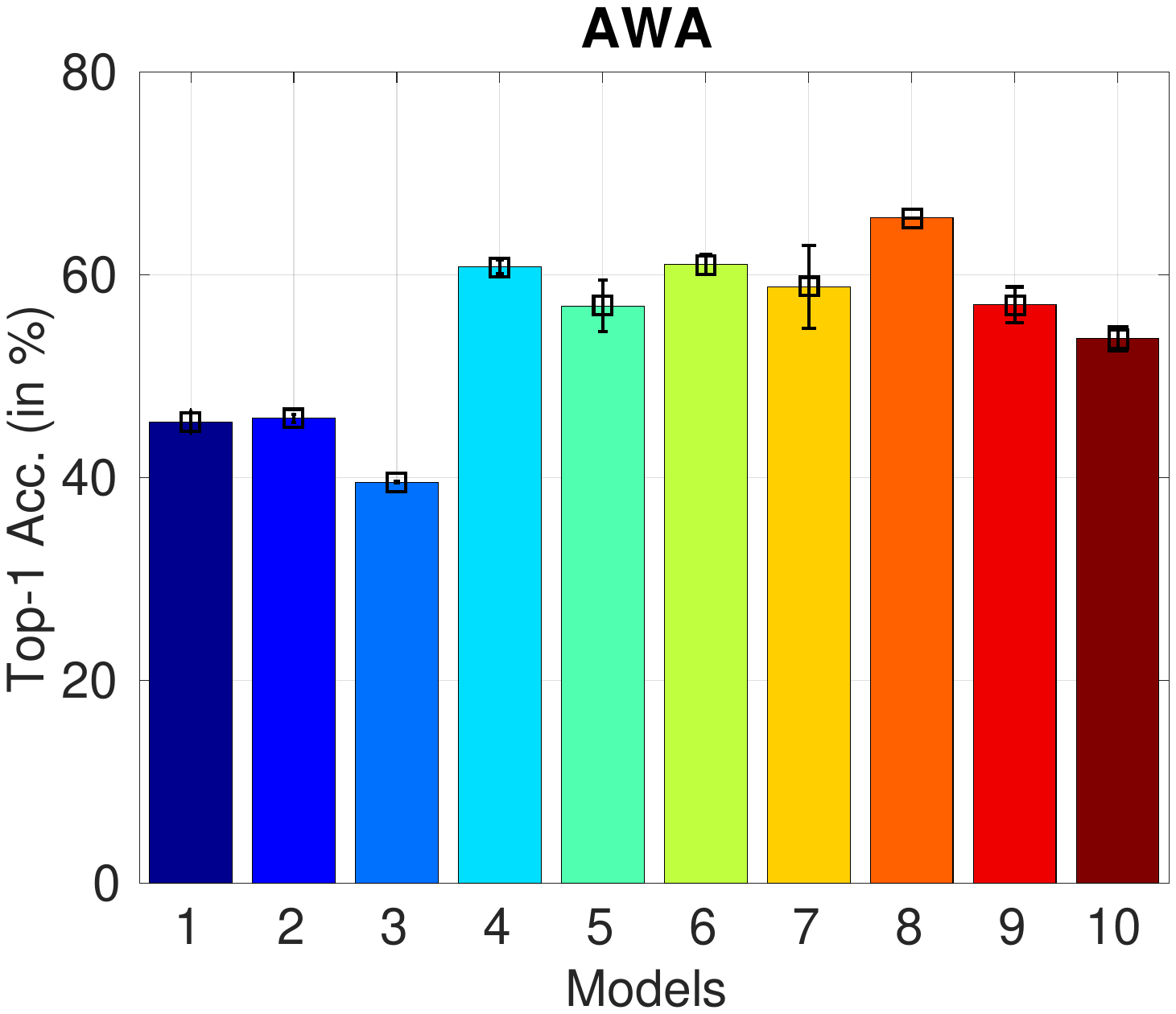}
    \includegraphics[width=0.23\linewidth, trim=20 5 70 10,clip]{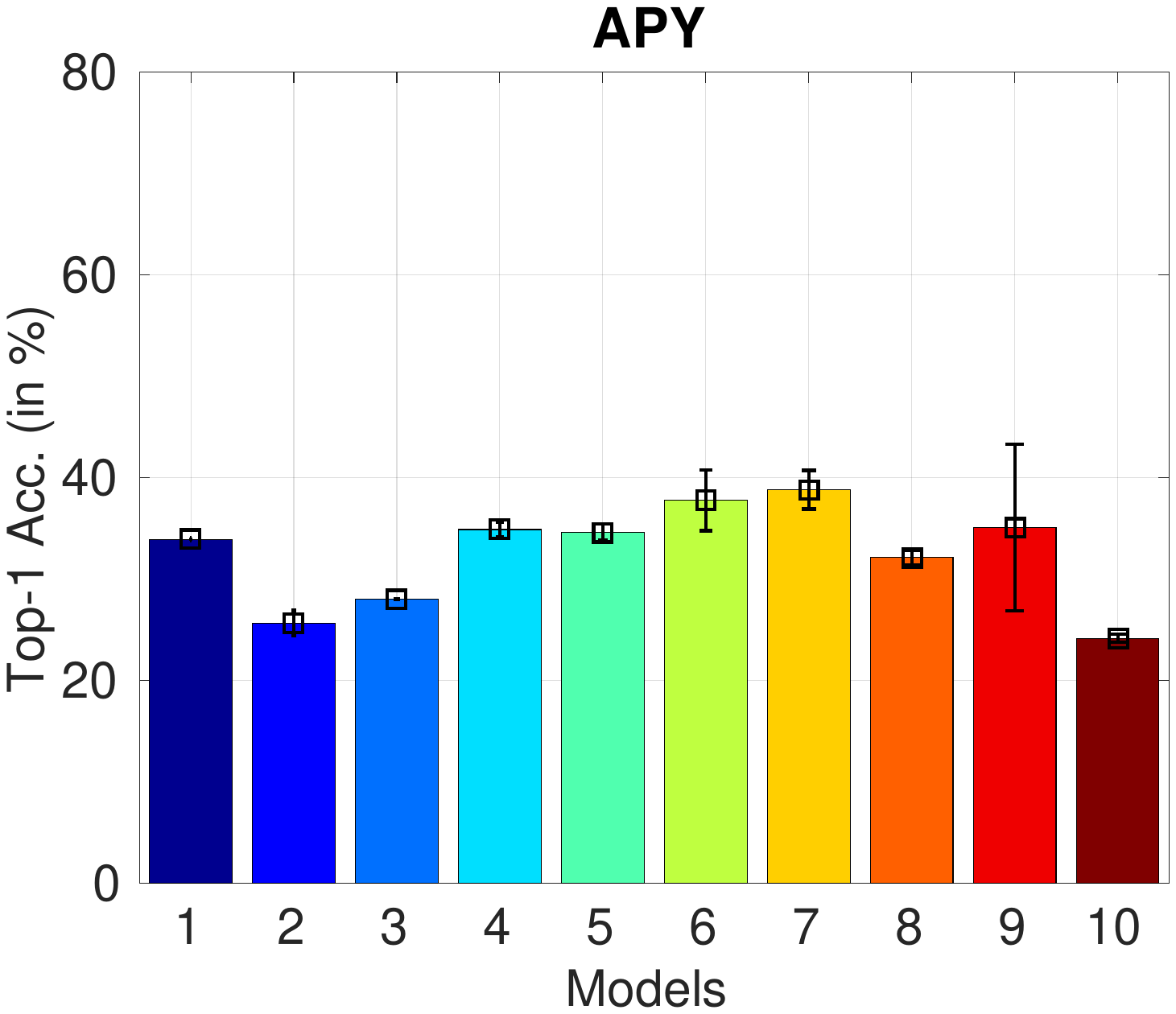}   
    \includegraphics[width=0.05\linewidth, trim=570 5 55 10,clip]{apy}
    \vspace{-1mm}
    \caption{Robustness of 10 methods evaluated on SUN, CUB, AWA, aPY using 3 validation set splits (results are on the same test split). Top: original split, Bottom: proposed split (Image embeddings = ResNet). We measure top-1 accuracy in \%.} 
    \vspace{-3mm}
	\label{fig:bar_plot}
\end{figure*}

\myparagraph{Reproduced Results vs Standard Split (SS).}
In addition to \cite{LNH13,ZV15,XASNHS16,ARWLS15,RT15,CCGS16,SGMN13}, we re-implement~\cite{NMBSSFCD14,FCSBDRM13,APHS15} based on the original publications. We use train, validation, test splits as provided in~\autoref{tab:datasets} and report results on~\autoref{tab:zeroshot} with deep ResNet features. DAP~\cite{LNH13} uses hand-crafted image features and thus reproduced results with those features are significantly lower than the results with deep features ($22.1\%$ vs $38.9\%$). When we investigate the results in detail, we noticed two irregularities with reported results on SUN. First, SSE~\cite{ZV15} and ESZSL~\cite{RT15} report results on a test split with $10$ classes whereas the standard split of SUN contains $72$ test classes ($74.5\%$ vs $54.5\%$ with SSE~\cite{ZV15} and $64.3\%$ vs $57.3\%$ with ESZSL~\cite{RT15}). Second, after careful examination and correspondence with the authors of SYNC~\cite{CCGS16}, we detected that SUN features were extracted with a MIT Places~\cite{zhou2014learning} pre-trained model. As MIT Places dataset intersects with both training and test classes of SUN dataset, it is expected to lead to significantly better results than ImageNet pre-trained model ($62.8\%$ vs $59.1\%$). 

\myparagraph{Results on Standard (SS) and Proposed Splits (PS).} We propose new dataset splits (see details in~\autoref{sec:datasets}) insuring that test classes do not belong to the ImageNet1K that is used to pre-train ResNet. We compare these results (PS) with previously published standard split (SS) results in~\autoref{tab:zeroshot}. Our first observation is that the results on PS is significantly lower than SS for AWA. This is expected as most of the test classes in SS is included in ImageNet 1K. On the other hand, for fine-grained datasets CUB and SUN, the results are not significantly effected.  Our second observation regarding the method ranking is as follows. On SS, SYNC~\cite{CCGS16} is the best performing method  on SUN ($59.1\%$) and aPY ($39.7\%$) datasets whereas SJE~\cite{ARWLS15} performs the best on CUB ($55.3\%$) and ALE~\cite{APHS15} performs the best on AWA ($78.6\%$) dataset. On PS, ALE~\cite{APHS15} performs the best on SUN ($58.1\%$) and aPY ($39.7\%$), SYNC~\cite{CCGS16} on CUB ($56.0\%$) and SJE~\cite{ARWLS15} on AWA ($65.6\%$). Note that ALE, SJE and DEVISE all use max-margin bi-linear compatibility learning framework. 

\myparagraph{Robustness.} We evaluate robustness of $10$ methods to parameters by setting them on $3$ different validation splits while keeping the test split intact. We report results on SS (\autoref{fig:bar_plot}, top) and PS (\autoref{fig:bar_plot}, bottom). On SUN and CUB, the results are stable across methods and across splits. This is expected as these datasets have balanced number of images across classes and due to their fine-grained nature, the validation splits are similar. On the other hand, AWA and aPY being small and coarse-grained datasets have several issues. First, many of the test classes on AWA and aPY are included in ImageNet1K. Second, they are not well balanced, i.e. different validation class splits contain significantly different number of images. Third, the class embeddings are far from each other, i.e. objects are semantically different, therefore different validation splits learn a different mapping between images and classes. 

\myparagraph{Visualizing Method Ranking.} We rank the $10$ methods based on their per-class top-1 accuracy using the non-parametric Friedman test~\cite{GH08}, which does not assume a distribution on performance but rather uses algorithm ranking. Each entry of the rank matrix on~\autoref{fig:rank_matrix} indicates the number of times the method is ranked at the first to tenth rank. We then compute the mean rank of each method and order them based on that. Our general observation is that the highest ranked method on the standard split (SS) is SYNC while on the proposed split (PS) it is ALE. These results indicate the importance of choosing zero-shot splits carefully. On the proposed split, the three highest ranked methods are compatibility learning methods, i.e. ALE, DEVISE and SJE whereas the three lowest ranked methods are attribute classifier learning or hybrid methods, i.e. DAP, CMT and CONSE. Therefore, max-margin compatibility learning methods lead to consistently better results in the zero-shot learning task compared to learning independent classifiers.    

{
\setlength{\tabcolsep}{2pt}
\renewcommand{\arraystretch}{1.3}
\begin{table}[t]
 \vspace{-3mm}
 \centering
 \resizebox{\linewidth}{!}{%
   \begin{tabular}{l c c |c c c |c c c |c}
     & \multicolumn{2}{c}{\textbf{Hierarchy}} & \multicolumn{3}{c}{\textbf{Most Populated}} & \multicolumn{3}{c}{\textbf{Least Populated}} & \textbf{All} \\
     \textbf{Method} & \textbf{2 H} & \textbf{3 H} & \textbf{500} & \textbf{1K} & \textbf{5K} & \textbf{500} & \textbf{1K} & \textbf{5K} & \textbf{20K} \\     
     \hline
     CONSE~\cite{NMBSSFCD14} & $7.63$ & $2.18$ & $12.33$ & $8.31$ & $3.22$ & $3.53$ & $2.69$ & $1.05$ & $0.95$ \\
     CMT~\cite{SGMN13} & $2.88$ & $0.67$ & $5.10$ & $3.04$ & $1.04$ & $1.87$ & $1.08$ & $0.33$ & $0.29$ \\
     LATEM~\cite{XASNHS16} & $5.45$ & $1.32$ & $10.81$ & $6.63$ & $1.90$ & $4.53$ & $2.74$ & $0.76$ & $0.50$ \\
      ALE~\cite{APHS15} & $5.38$ & $1.32$ & $10.40$ & $6.77$ & $2.00$ & $4.27$ & $2.85$ & $0.79$& $0.50$ \\
     DEVISE~\cite{FCSBDRM13} & $5.25$ & $1.29$ & $10.36$ & $6.68$ & $1.94$ & $4.23$ & $2.86$ & $0.78$ & $0.49$ \\
     SJE~\cite{ARWLS15} & $5.31$ & $1.33$ & $9.88$ & $6.53$ & $1.99$ & $4.93$ & $2.93$ & $0.78$ & $0.52$ \\   
     ESZSL~\cite{RT15} & $6.35$ & $1.51$ & $11.91$ & $7.69$ & $2.34$ & $4.50$ & $3.23$ & $0.94$ & $0.62$ \\
     SYNC~\cite{CCGS16} & $\mathbf{9.26}$ & $\mathbf{2.29}$ & $\mathbf{15.83}$ & $\mathbf{10.75}$ & $\mathbf{3.42}$ & $\mathbf{5.83}$ & $\mathbf{3.52}$ & $\mathbf{1.26}$ & $\mathbf{0.96}$ \\
     \hline
   \end{tabular} 
   }
   \vspace{-4mm}
\caption{ImageNet with different splits: 2/3 H = classes with 2/3 hops away from 1K $\mathcal{Y}^{tr}$, 500/1K/5K most populated classes, 500/1K/5K least populated classes, All=20K categories of ImageNet. We measure top-1 accuracy in \%.}
\vspace{-3mm}
\label{tab:ImageNet}
\end{table}
}

\myparagraph{Results on ImageNet.} 
ImageNet scales the methods to a truly large-scale setting, thus these experiments provide further insights on how to tackle the zero-shot learning problem from the practical point of view. Here, we evaluate $8$ methods. We exclude DAP as attributes are not available for all ImageNet classes and SSE due to  scalability issues of the public implementation of the method. \autoref{tab:ImageNet} shows that the best performing method is SYNC~\cite{CCGS16} which may indicate that it performs well in large-scale setting or it can learn under uncertainty due to usage of Word2Vec instead of attributes. Another possibility is Word2Vec may be tuned for SYNC as it is provided by the same authors however making a strong claim requires a full evaluation on class embeddings which is out of the scope of this paper. Our general observation from all the methods is that in the most populated classes, the results are higher than the least populated classes which indicates that fine-grained subsets are more difficult. We consistently observe a large drop in accuracy between 1K and 5K most populated classes which is expected as 5K contains $\approx 6.6$M images, making the problem much more difficult than 1K ($\approx 1624$ images). On the other hand, All 20K results are poor for all methods which indicates the difficulty of this problem where there is a large room for improvement.

{
\renewcommand{\arraystretch}{1.1}
\begin{table*}[t]
 \vspace{-4mm}
 \centering
   \begin{tabular}{l c c c |c c c |c c c |c c c }
      & \multicolumn{3}{c}{\textbf{SUN}} & \multicolumn{3}{c}{\textbf{CUB}} & \multicolumn{3}{c}{\textbf{AWA}} & \multicolumn{3}{c}{\textbf{aPY}}     \\
     \textbf{Method} &  \textbf{ts} & \textbf{tr} & \textbf{H} & \textbf{ts} & \textbf{tr} & \textbf{H} & \textbf{ts} & \textbf{tr} & \textbf{H} & \textbf{ts} & \textbf{tr} & \textbf{H}\\
     \hline
     DAP~\cite{LNH13} & $4.2$ &$25.1$ & $7.2$ & $1.7$ & $67.9$ & $3.3$ & $0.0$ & $88.7$ & $0.0$ & $4.8$ & $78.3$ & $9.0$   \\
     CONSE~\cite{NMBSSFCD14} & $6.8$ & $35.9$ & $11.4$ & $2.0$ & $\mathbf{70.6}$ & $3.9$ & $0.4$ & $\mathbf{89.6}$ & $0.8$ & $0.0$ & $\mathbf{91.2}$ & $0.0$   \\
     CMT~\cite{SGMN13} & $8.1$ & $21.8$ & $11.8$ & $7.2$ & $49.8$ & $12.6$ & $0.9$ & $87.6$ & $1.8$ & $1.4$ & $85.2$ & $2.8$   \\
     CMT*~\cite{SGMN13} & $8.7$ & $28.0$ & $13.3$ & $4.7$ & $60.1$ & $8.7$ & $8.4$ & $86.9$ & $15.3$ & $\mathbf{10.9}$ & $74.2$ & $\mathbf{19.0}$   \\   
     SSE~\cite{ZV15} & $2.1$ & $36.4$ & $4.0$ & $8.5$ & $46.9$ & $14.4$ & $7.0$ & $80.5$ & $12.9$ & $0.3$ & $78.4$ & $0.6$   \\
     LATEM~\cite{XASNHS16} & $14.7$ & $28.8$ & $19.5$ & $15.2$ & $57.3$ & $24.0$ & $7.3$ & $71.7$ & $13.3$ & $1.3$ & $71.4$ & $2.6$   \\
     ALE~\cite{APHS15}  & $\mathbf{21.8}$ & $33.1$ & $\mathbf{26.3}$ & $23.7$ & $62.8$ & $\mathbf{34.4}$ & $\mathbf{16.8}$ & $76.1$ & $\mathbf{27.5}$ & $4.6$ & $73.7$ & $8.7$ \\
     DEVISE~\cite{FCSBDRM13}  & $16.9$ & $27.4$ & $20.9$ & $\mathbf{23.8}$ & $53.0$ & $32.8$ & $13.4$ & $68.7$ & $22.4$ & $3.5$ & $78.4$ & $6.7$   \\
     SJE~\cite{ARWLS15} & $14.4$ & $29.7$ & $19.4$ & $23.5$ & $59.2$ & $33.6$ & $11.3$ & $74.6$ & $19.6$ & $1.3$ & $71.4$ & $2.6$   \\     
     ESZSL~\cite{RT15} & $11.0$ & $27.9$ & $15.8$ & $14.7$ & $56.5$ & $23.3$ & $6.6$ & $75.6$ & $12.1$ & $2.4$ & $70.1$ & $4.6$   \\
     SYNC~\cite{CCGS16} & $7.9$ & $\mathbf{43.3}$ & $13.4$ & $11.5$  & $70.9$ & $19.8$ & $9.0$ & $88.9$ & $16.3$ & $7.4$ & $66.3$ & $13.3$ \\
     \hline
   \end{tabular} 
   \vspace{-2mm}
\caption{Generalized Zero-Shot Learning on Proposed Split Version 2.0 (PS) measuring ts = Top-1 accuracy on $\mathcal{Y}^{ts}$, tr=Top-1 accuracy on $\mathcal{Y}^{tr + ts}$), H = harmonic mean (CMT*: CMT with novelty detection). We measure top-1 accuracy in \%.}
\vspace{-3mm}
\label{tab:openset}
\end{table*}
}

\subsection{Generalized Zero-Shot Learning Results}
In real world applications, image classification systems do not have access to whether a novel image belongs to a seen or unseen class in advance. Hence, generalized zero-shot learning is interesting from a practical point of view. Here, we use same models trained on zero-shot learning setting on our proposed splits (PS). We evaluate performance on both $\mathcal{Y}^{tr}$ and $\mathcal{Y}^{ts}$, i.e. using held-out images from $\mathcal{Y}^{ts}$.

As shown in \autoref{tab:openset}, generalized zero-shot results are significantly lower than zero-shot results as training classes are included in the search space. Another interesting observation is that compatibility learning frameworks, e.g. ALE, DEVISE, SJE, perform well on test classes. However, methods that learn independent attribute or object classifiers, e.g. DAP and CONSE, perform well on training classes. Due to this discrepancy, we evaluate the harmonic mean which takes a weighted average of training and test class accuracy. H measure ranks ALE as the best performing method on SUN, CUB and AWA datasets whereas on aPY dataset CMT* performs the best. Note that CMT* has an integrated novelty detection phase for which the method receives another  supervision signal determining if the image belongs to a train or a test class. As a summary, generalized zero-shot learning setting provides one more level of detail on the performance of zero-shot learning methods. Our take-home message is that the accuracy of training classes is as important as the accuracy of test classes in real world scenarios. Therefore, methods should be designed in a way that they are able to predict labels well in train and test classes.  
\begin{figure}[t]
	\centering 
    \includegraphics[width=0.49\columnwidth, trim=0 60 70 0,clip]{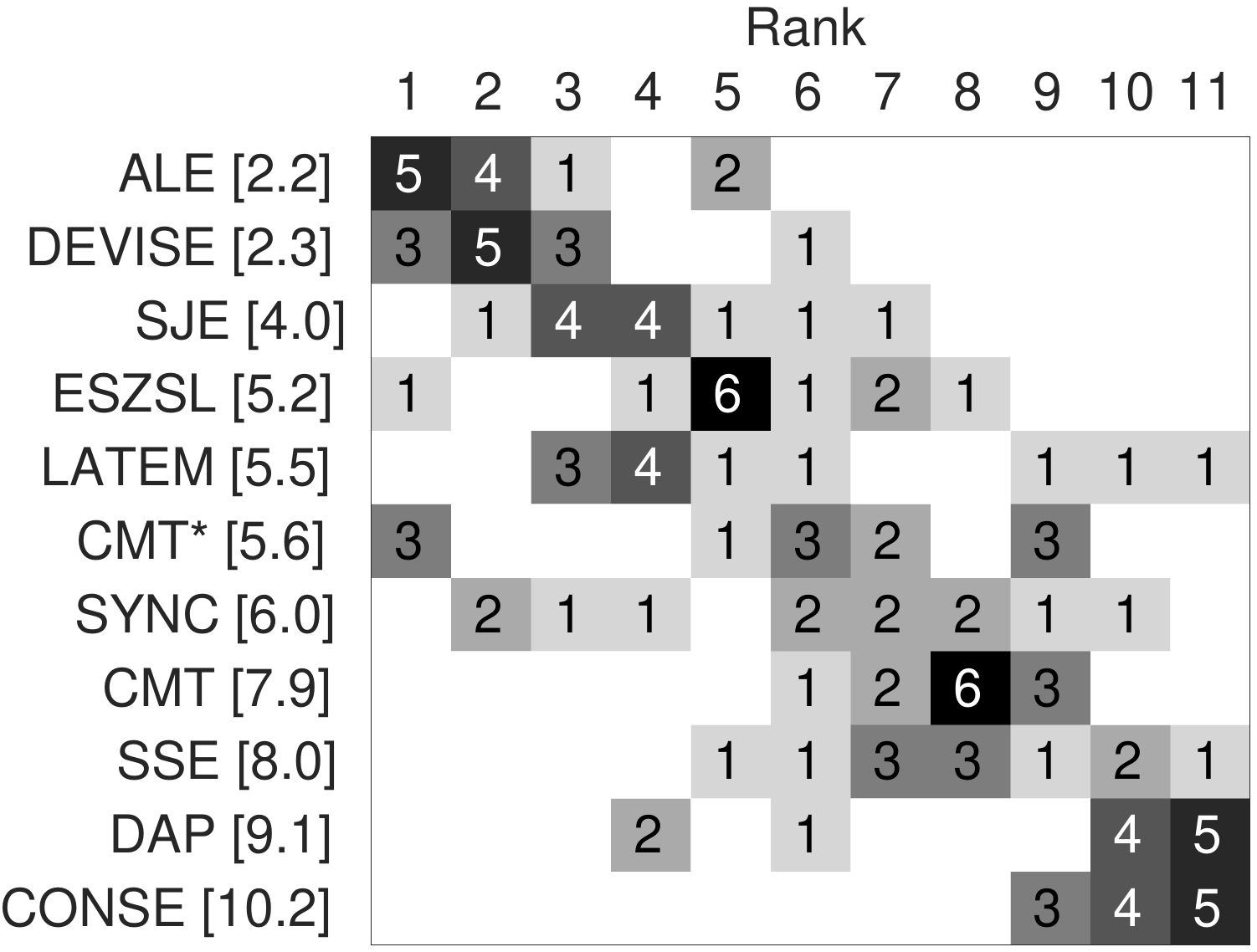}
	\includegraphics[width=0.49\columnwidth, trim=0 60 70 0,clip]{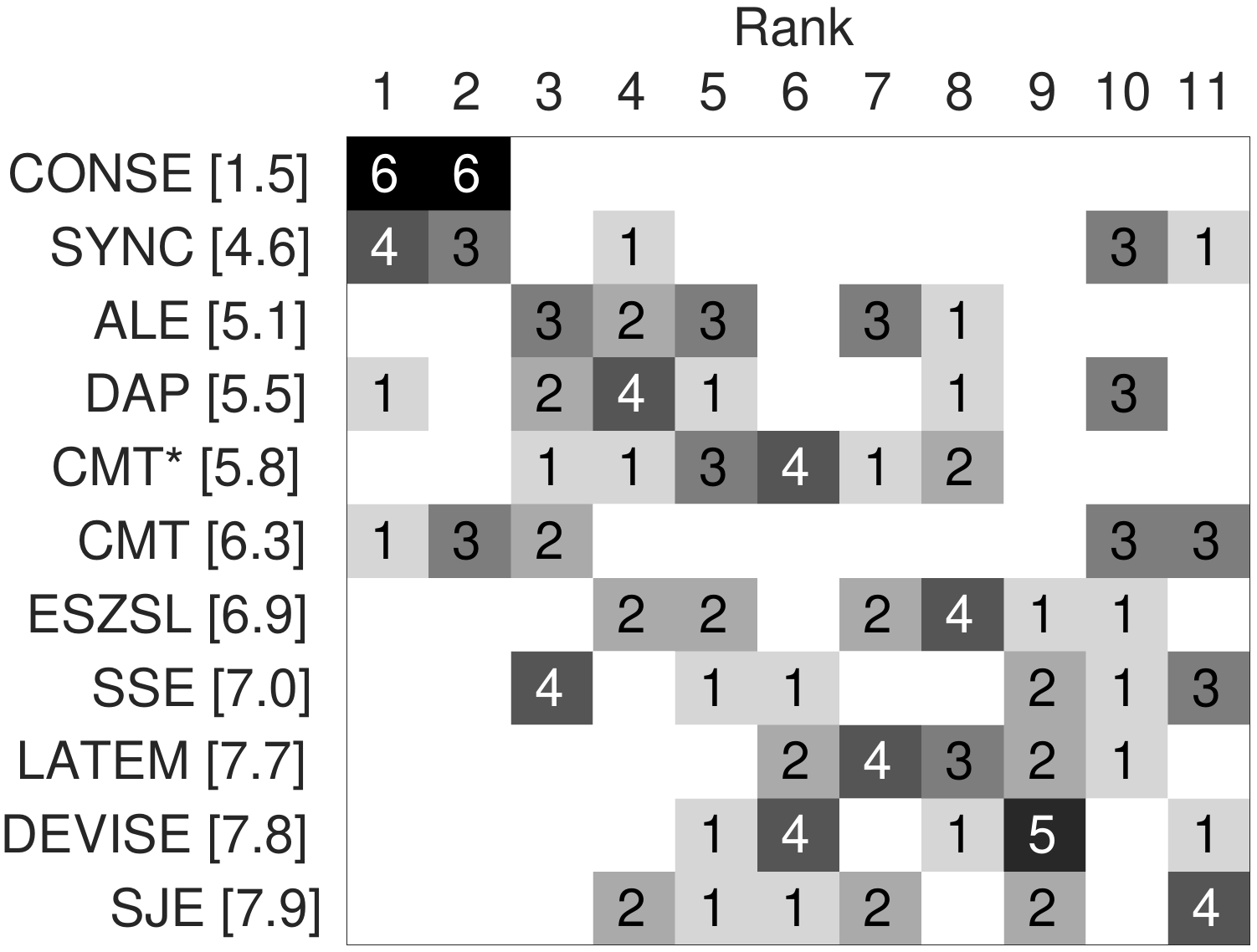}
    \includegraphics[width=0.49\columnwidth, trim=0 60 70 0,clip]{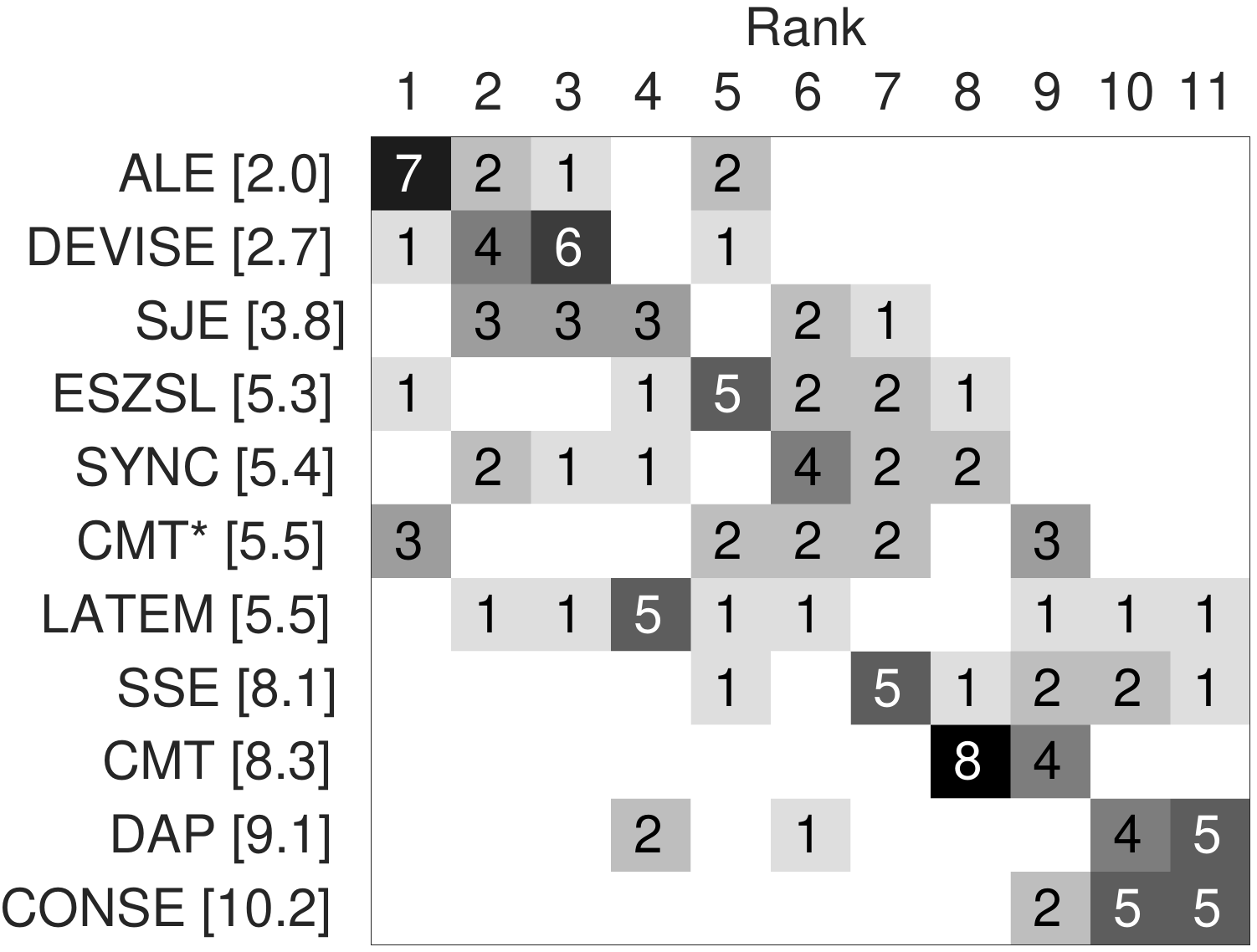}
    \vspace{-2mm}
    \caption{Ranking 11 models on the proposed split (PS) in generalized zero-shot learning setting. Top-Left: on unseen cla sses (ts) accuracy, Top-Right: on seen classes (tr) accuracy,  Bottom: on Harmonic mean (H).} \vspace{-3mm} 
	\label{fig:rankGeneralized}
\end{figure}

\myparagraph{Visualizing Method Ranking.} Similar to the analysis in the previous section, we rank the $11$ methods based on per-class top-1 accuracy on train classes, test classes and based on Harmonic mean of the two. Looking at the rank matrix obtained by evaluating on test classes, i.e. \autoref{fig:rankGeneralized} top left, highest ranked 5 methods are the same as in~\autoref{fig:rank_matrix}, i.e. ALE, DEVISE, SJE, LATEM, ESZSL while overall the absolute numbers are lower. Looking at the rank matrix obtained by evaluating the harmonic mean, i.e. \autoref{fig:rankGeneralized} bottom, the highest ranked 3 methods are the same as in~\autoref{fig:rank_matrix}, i.e. ALE, DEVISE, SJE. Looking at the rank matrix obtained by evaluating on train classes, i.e. \autoref{fig:rankGeneralized} top right, our observations are different from~\autoref{fig:rank_matrix}. ALE is ranked the 3rd but other highest ranked methods are at the bottom of this rank list. These results clearly suggest that we should not only optimize for test class accuracy but also for train class accuracy when evaluating zero-shot learning. Our final observation from~\autoref{fig:rankGeneralized} is that CMT* is better than CMT in all cases which supports the argument that a simple novelty detection scheme helps to improve results.

\section{Conclusion}
\label{sec:conc}
In this work, we evaluated a significant number of state-of-the-art zero-shot learning methods on several datasets within a unified evaluation protocol both in zero-shot and generalized zero-shot settings. Our evaluation showed that compatibility learning frameworks have an edge over learning independent object or attribute classifiers and also over hybrid models. We discovered that some standard zero-shot splits may treat feature learning disjoint from the training stage and accordingly proposed new dataset splits. Moreover, disjoint training and validation class split is a necessary component of parameter tuning in zero-shot learning setting. Including training classes in the search space while evaluating the methods, i.e. generalized zero-shot learning, provides an interesting playground for future research. In summary, our work extensively evaluated the good and bad aspects of zero-shot learning while sanitizing the ugly ones.

{\small
\bibliographystyle{ieee}
\bibliography{egbib}
}

\end{document}


\title{Supplementary Material:\\ Zero-Shot Learning - The Good, the Bad and the Ugly}

\author{First Author\\
Institution1\\
Institution1 address\\
{\tt\small firstauthor@i1.org}
\and
Second Author\\
Institution2\\
First line of institution2 address\\
{\tt\small secondauthor@i2.org}
}
\newcommand{\edit}[1]{\textcolor{red}{\textbf{Modified: #1}}}
\newcommand{\add}[1]{\textcolor{red}{\textbf{Added: #1}}}
\newcommand{\toAdd}[1]{\textcolor{red}{\textbf{To add #1?}}}
\maketitle
\myparagraph{Reproducing results.}
In this section, we extend our reproduced results and original results in Tab.\ref{tab:reproduce} for 6 models which have both codes and data available to CUB and aPY, i.e. SUN and CUB results appear in the main paper and we include these results here for completion. We obtain identical results as DAP, SJE and SYNC report in their original paper. The slightly different results of LATEM could be explained by its non-convexity and thus the sensibility to initialization. ESZSL picks a validation set randomly at each run which leads to slightly different hyperparameters and results. Regarding to SSE, we get closed results on aPY because the hyperparameters of aPY are provided in the demo code, by hyperparameter tuning through cross-validation we get significantly better results on CUB ($44.2\%$ vs $30.4\%$).

\begin{table}[h]
 \centering
   \begin{tabular}{l c c c c c c c c}
     & \multicolumn{8}{c }{\textbf{Accuracy (Top1 in \%)}} \\
& \multicolumn{2}{c }{\textbf{SUN}} & \multicolumn{2}{c }{\textbf{CUB}} & \multicolumn{2}{c }{\textbf{AWA}} & \multicolumn{2}{c }{\textbf{aPY}} \\
      \textbf{Model} & \textbf{R} & \textbf{O}   & \textbf{R} & \textbf{O} & \textbf{R} & \textbf{O} & \textbf{R} & \textbf{O}\\     
     \hline
     DAP & $22.1$ & $22.2$ & $-$ & $-$ & $41.4$ & $41.4$ & $19.1$ & $19.1$\\
     SSE & $83.0$ & $82.5$ & $44.2$ & $30.4$ & $64.9$ & $76.3$ & $45.7$ & $46.2$ \\
     LATEM & $-$ & $-$ & $45.1$ & $45.5$ & $71.2$ & $71.9$ & $-$ & $-$\\
     SJE & $-$ & $-$ & $50.1$ & $50.1$ & $67.2$ & $66.7$ & $-$ & $-$\\
     ESZSL & $64.3$ & $65.8$ & -- & -- & $48.0$ & $49.3$ & $14.3$ & $15.1$\\
     SYNC & $62.8$ & $62.8$ & $53.4$ & $53.4$ & $69.7$ & $69.7$ & -- & --\\ 
     \hline
   \end{tabular} 
\caption{Reproducing zero-shot learning results: O = Original results published in the paper, R = Reproduced using provided image features and code (-- : this paper did not report performance on this dataset). Along with the results obtained on CUB and aPY, for completion we also report results on SUN and AWA (same as our main paper). }
\label{tab:reproduce}
\end{table}

\myparagraph{GoogleNet vs ResNet.} Our image embeddings presented in the main paper are $2048$-dim top-layer pooling units of the $101$-layered ResNet. Here we show that it performs better than $1,024$-dim top-layer pooling units of GoogleNet. Both networks are pre-trained on ImageNet 1K and not fine-tuned. We present zero-shot learning results on Standard Split (SS) using GoogleNet vs ResNet features on~\autoref{tab:zeroshot}. We observe results improving consistently and significantly in all cases for SUN, CUB, AWA datasets and almost all cases for aPY dataset with ResNet features. 

\begin{table}[h]
 \centering
   \begin{tabular}{l c c c c c c c c}
     & \multicolumn{2}{c }{\textbf{SUN}} & \multicolumn{2}{c }{\textbf{CUB}} & \multicolumn{2}{c }{\textbf{AWA}} & \multicolumn{2}{c }{\textbf{aPY}}\\     
      \textbf{Model} & \textbf{G} & \textbf{R}   & \textbf{G} & \textbf{R} & \textbf{G} & \textbf{R} & \textbf{G} & \textbf{R}\\
     \hline
     DAP  & $34.0$ & $38.9$ & $35.6$ & $37.5$ & $54.7$ & $57.1$ & $26.4$ & $35.2$ \\
     CONSE & $37.0$ & $44.2$ & $35.1$ & $36.7$ & $57.8$ & $63.6$ & $23.1$ & $25.9$\\
     CMT & $34.7$ & $41.9$ & $29.1$ & $37.3$ & $56.8$ & $58.9$ & $22.0$ & $26.9$ \\
     SSE & $46.7$ & $54.5$ & $42.1$ & $43.7$ & $64.8$ & $68.8$ & $32.4$ & $31.1$ \\
     LATEM & $48.8$ & $56.9$ & $45.5$ & $49.4$ & $71.9$ & $74.8$ & $29.7$ & $34.5$ \\
     ALE & $50.7$ & $59.1$ & $49.0$ & $53.2$ & $76.5$ & $78.6$ & $30.9$ & $30.9$  \\
     DEVISE  & $50.1$ & $57.5$ & $48.2$ & $53.2$ & $73.4$ & $72.9$ & $31.5$ & $35.4$ \\
     SJE & $45.9$ & $57.1$ &  $50.1$ & $55.3$ & $66.7$ & $76.7$ & $30.1$ & $32.0$ \\
     ESZSL  & $48.1$ & $57.3$ & $51.2$ & $55.1$ & $70.8$ & $74.7$ & $32.2$ & $34.4$\\
     SYNC & $49.0$ & $59.1$ & $51.6$ & $54.1$ & $69.3$ & $72.2$ & $40.1$ & $39.7$  \\ 
     \hline
   \end{tabular} 
\caption{Zero-shot learning results on Standard Split (SS) using GoogleNet vs ResNet features. We observe results improving consistently in all cases with ResNet features. This is our motivation of reporting results on ResNet features in the main paper.}
\label{tab:zeroshot}
\end{table}

%

 \myparagraph{Top-1/5/10 Accuracy on ImageNet: Zero-shot Learning.}
 Top-K accuracy is defined as the percentage of test images whose truth labels appear in the top-K predictions. Here we first compute the top-K accuracy for each class and take the average over all classes. In Tab.~\ref{tab:imagenet_zsl}, we observe that SYNC achieves the best top-1/5/10 accuracies for all the test sets of imageNet in zero-shot learning setting. If we consider the top-1 accuracy, CONSE ranks the second place except on \textbf{L 500} and \textbf{L 1K}. But if we look at the top-5/10 accuracies, ESZSL outperforms CONSE in many cases, for instance, top-5 accuracies on \textbf{2 H} are ESZSL ($20.22\%$) vs CONSE ($19.46\%$), and top-10 accuracies on \textbf{2 H} are ESZSL ($30.22\%$) vs CONSE ($27.24\%$).

\begin{table}[h]
 \centering
 \setlength{\tabcolsep}{3pt}
\renewcommand{\arraystretch}{1.2}
\begin{tabular}{ l |c c c| c c c| c c c| c c c| c c c|}
  Method & \multicolumn{3}{c}{\textbf{2 H}} & \multicolumn{3}{c}{\textbf{3 H}} & \multicolumn{3}{c}{\textbf{M 500}} & \multicolumn{3}{c}{\textbf{M 1K}} & \multicolumn{3}{c}{\textbf{M 5K}} \\ 
  & Top-1 & Top-5 & Top-10 & Top-1 & Top-5 & Top-10 & Top-1 & Top-5 & Top-10 & Top-1 & Top-5 & Top-10 & Top-1 & Top-5 & Top-10 \\
   \hline
  CONSE & 7.63 & 19.46 & 27.24 & 2.18 & 6.01 & 8.84 & 12.33 & 27.74 & 37.55 & 8.31 & 19.48 & 27.22 & 3.22 & 8.44 & 12.26 \\
  CMT & 2.88 & 7.48 & 10.77 & 0.67 & 2.17 & 3.36 & 5.10 & 12.94 & 18.07 & 3.04 & 8.46 & 12.37 & 1.04 & 3.12 & 4.88 \\
  LATEM & 5.47 & 17.82 & 27.08 & 1.31 & 4.65 & 7.68 & 10.80 & 30.41 & 42.49 & 6.61 & 21.10 & 30.79 & 1.90 & 6.80 & 11.15 \\
  ALE & 5.38 & 17.67 & 26.86 & 1.32 & 4.60 & 7.49 & 10.40 & 29.61 & 41.58 & 6.77 & 20.59 & 30.01 & 2.00 & 6.97 & 11.25 \\
  DEVISE & 5.28 & 17.47 & 26.59 & 1.30 & 4.52 & 7.35 & 10.39 & 29.39 & 41.23 & 6.69 & 20.45 & 29.80 & 1.95 & 6.81 & 11.03\\
  SJE & 5.31 & 16.94 & 26.67 & 1.33 & 4.45 & 7.26 & 9.88 & 28.74 & 40.06 & 6.53 & 19.61 & 28.72 & 1.99 & 6.73 & 10.78 \\
  ESZSL & 6.35 & 20.22 & 30.22 & 1.51 & 5.26 & 8.61 & 11.91 & 32.98 & 45.77 & 7.69 & 23.11 & 33.39 & 2.34 & 7.88 & 12.75 \\
  SYNC & \textbf{9.26} & \textbf{26.35} & \textbf{36.96} & \textbf{2.29} & \textbf{7.55} & \textbf{11.86} & \textbf{15.83} & \textbf{38.82} & \textbf{51.59} & \textbf{10.74} & \textbf{28.30} & \textbf{39.01} & \textbf{3.44} & \textbf{10.86} & \textbf{16.68} \\
\hline
  Method & \multicolumn{3}{c}{\textbf{L 500}} & \multicolumn{3}{c}{\textbf{L 1K}} & \multicolumn{3}{c}{\textbf{L 5K}} & \multicolumn{3}{c}{\textbf{All}} \\
  & Top-1 & Top-5 & Top-10 & Top-1 & Top-5 & Top-10 & Top-1 & Top-5 & Top-10 & Top-1 & Top-5 & Top-10 & \\
  CONSE & 3.53 & 12.17 & 17.93 & 2.69 & 7.42 & 11.72 & 1.05 & 3.00 & 4.78 & 0.95 & 2.68 & 4.05 \\
  CMT &  1.87 & 4.37 & 5.80 & 1.08 & 3.08 & 4.01 & 0.33 & 1.19 & 2.05 & 0.29 & 0.93 & 1.46\\
  LATEM & 4.70 & 13.93 & 21.00 & 2.84 & 8.52 & 13.82 & 0.77 & 2.98 & 4.87 & 0.50 & 1.80 & 3.09 \\
  ALE & 4.27 & 13.70 & 20.30 & 2.85 & 8.34 & 13.44 & 0.79 & 2.94 & 4.72 & 0.50 & 1.83 & 3.11\\
  DEVISE & 4.63 & 12.37 & 20.43 & 2.94 & 8.36 & 13.04 & 0.79 & 2.90 & 4.62 & 0.49 & 1.79 & 3.05\\
  SJE &  4.93 & 11.90 & 19.77 & 2.93 & 8.19 & 12.78 & 0.78 & 2.77 & 4.45 & 0.52 & 1.84 & 3.04\\
  ESZSL & 4.50 & 13.83 & 21.57 & 3.23 & 9.25 & 14.45 & 0.94 & 3.42 & 5.53 & 0.62 & 2.14 & 3.61\\
  SYNC & \textbf{5.83} & \textbf{16.53} & \textbf{25.27} & \textbf{3.52} & \textbf{11.45} & \textbf{17.56} & \textbf{1.25} & \textbf{4.39} & \textbf{7.02} & \textbf{0.96} & \textbf{3.22} & \textbf{5.19}\\
  \cline{1-13}
\end{tabular}
\caption{Top-1/5/10 accuracy in zero-shot learning setting on ImageNet with different test sets: 2/3 H=all classes two/three hops away from 1K training classes, M 500/1K/5K= 500/1K/5K most populated classes, L 500/1K/5K= 500/1K/5K least populated classes, All=20K categories of ImageNet.}
\label{tab:imagenet_zsl}
\end{table}

\myparagraph{Top-1/5/10 Accuracy on ImageNet: Generalized Zero-shot Learning.} In this setting, we add the 1K training classes into the test sets as distractors at the evaluation time. The results in Tab.~\ref{tab:imagenet_gzsl} show that the performance ranking of models has totally changed comparing with zero-shot learning setting. The top-1 accuracies of both SYNC and CONSE are close to $0$ for all test sets. ALE achieves the best accuracies in 11 cases out of 27 cases, which is the best performing model among all. CONSE starts to recover from top-5/10 accuracies and obtains 7 best accuracies our of 27 cases, which is the second best model.   

\begin{table}[t]
 \centering
 \setlength{\tabcolsep}{3pt}
\renewcommand{\arraystretch}{1.2}
\begin{tabular}{ l |c c c| c c c| c c c| c c c| c c c| c c c| }
  Method & \multicolumn{3}{c}{\textbf{2 H}} & \multicolumn{3}{c}{\textbf{3 H}} & \multicolumn{3}{c}{\textbf{M 500}} & \multicolumn{3}{c}{\textbf{M 1K}} & \multicolumn{3}{c}{\textbf{M 5K}} \\ 
  & Top-1 & Top-5 & Top-10 & Top-1 & Top-5 & Top-10 & Top-1 & Top-5 & Top-10 & Top-1 & Top-5 & Top-10 & Top-1 & Top-5 & Top-10\\
   \hline
  CONSE & 0.00 & 0.67 & 1.01 & 0.01 & \textbf{4.34} & \textbf{7.15} & 0.04 & 14.83 & 23.59 & 0.03 & \textbf{11.78} & 18.39 & 0.02 & \textbf{6.20} & \textbf{9.95} \\
  CMT & 1.10 & 5.05 & 7.83 & 0.34 & 1.66 & 2.76 & 1.60 & 6.24 & 9.63 & 1.19 & 4.77 & 7.72 & 0.56 & 2.37 & 3.93 \\
  LATEM & 2.01 & 9.55 & 17.02 & 0.75 & 3.56 & 6.27 & 2.64 & 13.59 & 23.67 & 2.09 & 10.64 & 18.73 & 1.02 & 5.02 & 8.77 \\
  ALE & \textbf{2.18} & 10.24 & 17.86 & \textbf{0.80} & 3.69 & 6.33 & 2.88 & \textbf{14.86} & 24.89 & \textbf{2.34} & 11.57 & 19.43 & \textbf{1.09} & 5.32 & 9.14 \\
  DEVISE & 2.15 & 10.16 & 17.57 & 0.78 & 3.61 & 6.21 & \textbf{2.90} & 14.68 & 24.70 & 2.33 & 11.39 & 19.20 & 1.08 & 5.20 & 8.95 \\
  SJE & 1.82 & \textbf{10.31} & 17.30 & 0.71 & 3.57 & 6.08 & 2.37 & 14.51 & 23.69 & 1.87 & 11.19 & 18.57 & 0.95 & 5.19 & 8.82 \\
  ESZSL & 1.34 & 10.29 & \textbf{19.33} & 0.54 & 3.70 & 6.78 & 1.68 & 14.68 & \textbf{26.55} & 1.50 & 11.32 & \textbf{20.65} & 0.78 & 5.37 & 9.74 \\
  SYNC & 0.00 & 6.10 & 14.58 & 0.01 & 2.34 & 5.59 & 0.00 & 7.35 & 16.39 & 0.00 & 6.14 & 13.86 & 0.00 & 3.55 & 7.92 \\
  \hline
  & \multicolumn{3}{c}{\textbf{L 500}} & \multicolumn{3}{c}{\textbf{L 1K}} & \multicolumn{3}{c}{\textbf{L 5K}} & \multicolumn{3}{c}{\textbf{All}} \\
  & Top-1 & Top-5 & Top-10 & Top-1 & Top-5 & Top-10 & Top-1 & Top-5 & Top-10 & Top-1 & Top-5 & Top-10 & \\
  CONSE & 0.00 & 0.00 & 0.00 & 0.00 & 0.00 & 0.00 & 0.00 & 0.36 & 0.71 & 0.00 & \textbf{2.01} & \textbf{3.39}\\
  CMT & 0.80 & 2.37 & 3.37 & 0.40 & 1.75 & 2.76 & 0.17 & 0.74 & 1.41 & 0.18 & 0.75 & 1.26\\
  LATEM  & 1.03 & 4.47 & 8.73 & 0.56 & 4.31 & 7.63 & 0.34 & 2.00 & 3.52 & 0.33 & 1.52 & 2.72\\
  ALE & \textbf{1.87} & 4.93 & \textbf{10.37} & 1.21 & \textbf{4.81} & \textbf{8.47} & \textbf{0.44} & 2.15 & 3.73 & \textbf{0.34} & 1.58 & 2.80\\
  DEVISE & 1.63 & 5.57 & 10.17 & 1.27 & 4.54 & 8.40 & \textbf{0.44} & \textbf{2.17} & 3.70 & 0.33 & 1.55 & 2.75\\
  SJE & 1.67 & \textbf{6.10} & 9.90 & \textbf{1.36} & 4.39 & 8.01 & 0.43 & 2.00 & 3.51 & 0.31 & 1.59 & 2.74\\
  ESZSL &  0.63 & 4.00 & 9.10 & 0.43 & 3.99 & 7.79 & 0.28 & 1.96 & \textbf{3.86} & 0.27 & 1.67 & 3.06\\
  SYNC & 0.00 & 0.70 & 2.87 & 0.00 & 0.71 & 2.48 & 0.00 & 0.56 & 1.78 & 0.00 & 1.18 & 2.69\\
  \cline{1-13}
\end{tabular}
\caption{Top-1/5/10 accuracy in generalized zero-shot learning setting on ImageNet with different test sets: 2/3 H=all classes two/three hops away from 1K training classes, M 500/1K/5K= 500/1K/5K most populated classes, L 500/1K/5K= 500/1K/5K least populated classes, All=20K categories of ImageNet.}
\label{tab:imagenet_gzsl}
\end{table}
